\documentclass[journal]{vgtc}                     

\onlineid{1206}

\vgtccategory{IEEE VIS Full Paper}

\vgtcpapertype{Applications}

\title{AortaDiff: Volume-Guided Conditional Diffusion Models for Multi-Branch Aortic Surface Generation}

\author{%
  \authororcid{Delin An}{0000-0002-7945-0201},
  \authororcid{Pan Du}{0000-0001-5445-1497},
  \authororcid{Jian-Xun Wang}{0000-0002-9030-1733}, and
  \authororcid{Chaoli Wang}{0000-0002-0859-3619}
}

\authorfooter{
  \item D.\ An and C.\ Wang are with 
the Department of Computer Science and Engineering, University of Notre Dame, Notre Dame, IN 46556, USA.\\
E-mail: \{dan3, chaoli.wang\}@nd.edu.
 \item P. Du is with the Department of Aerospace and Mechanical Engineering, University of Notre Dame, Notre Dame, IN 46556, USA.\\
E-mail: pdu@nd.edu.
 \item J.-X. Wang is with the Sibley School of Mechanical and Aerospace Engineering, Cornell University, Ithaca, NY 14850, USA.\\
E-mail: jw2837@cornell.edu. 
}

\abstract{
Accurate 3D aortic construction is crucial for clinical diagnosis, preoperative planning, and computational fluid dynamics (CFD) simulations, as it enables the estimation of critical hemodynamic parameters such as blood flow velocity, pressure distribution, and wall shear stress. Existing construction methods often rely on large annotated training datasets and extensive manual intervention. \pin{While the resulting meshes can serve for visualization purposes}, they struggle to produce geometrically consistent, well-constructed surfaces suitable for downstream CFD analysis. To address these challenges, we introduce AortaDiff, a diffusion-based framework that generates smooth aortic surfaces directly from CT/MRI volumes. AortaDiff first employs a volume-guided conditional diffusion model (CDM) to iteratively generate aortic centerlines conditioned on volumetric medical images. Each centerline point is then automatically used as a prompt to extract the corresponding vessel contour, ensuring accurate boundary delineation. Finally, the extracted contours are fitted into a smooth 3D surface, yielding a continuous, CFD-compatible mesh representation. AortaDiff offers distinct advantages over existing methods, including an end-to-end workflow, minimal dependency on large labeled datasets, and the ability to generate CFD-compatible aorta meshes with high geometric fidelity. \hot{Experimental results demonstrate that AortaDiff performs effectively even with limited training data, successfully constructing both normal and pathologically altered aorta meshes, including cases with aneurysms or coarctation. This capability enables the generation of high-quality visualizations and positions AortaDiff as a practical solution for cardiovascular research.}
}


\graphicspath{{figs/}{figures/}{pictures/}{images/}{./}} 

\usepackage{tabu}                      
\usepackage{booktabs}                  
\usepackage{lipsum}                    
\usepackage{mwe}                       

\usepackage{mathptmx}                  

\usepackage{graphicx}
\usepackage{times}
\usepackage[ruled,vlined]{algorithm2e}
\usepackage{amsfonts}
\usepackage{booktabs}         
\usepackage{gensymb}
\usepackage{amsmath}
\usepackage{comment}
\usepackage{times} 
\usepackage{caption}
\usepackage{bm}
\usepackage{multirow}
\usepackage{color} 
\usepackage{anyfontsize}
\usepackage{soul}

\usepackage{arydshln}
\usepackage{array}

\DeclareMathOperator*{\twod}{2D}
\DeclareMathOperator*{\threed}{3D}
\DeclareMathOperator*{\cf}{CurveFitting}

\newcommand{\hot}[1]{{\color{black} #1}}
\newcommand{\pin}[1]{{\color{black} #1}}
\newcommand{\delin}[1]{{\color{black} #1}}

\newenvironment{myitemize}{
\begin{itemize}
 \setlength{\itemsep}{1pt}
 \setlength{\parskip}{0pt}
 \setlength{\parsep}{0pt}}{\end{itemize}
 
}

\begin{document}

\firstsection{Introduction}
\maketitle

\begin{figure}[htb]
    \centering
    \includegraphics[width=0.75\linewidth]{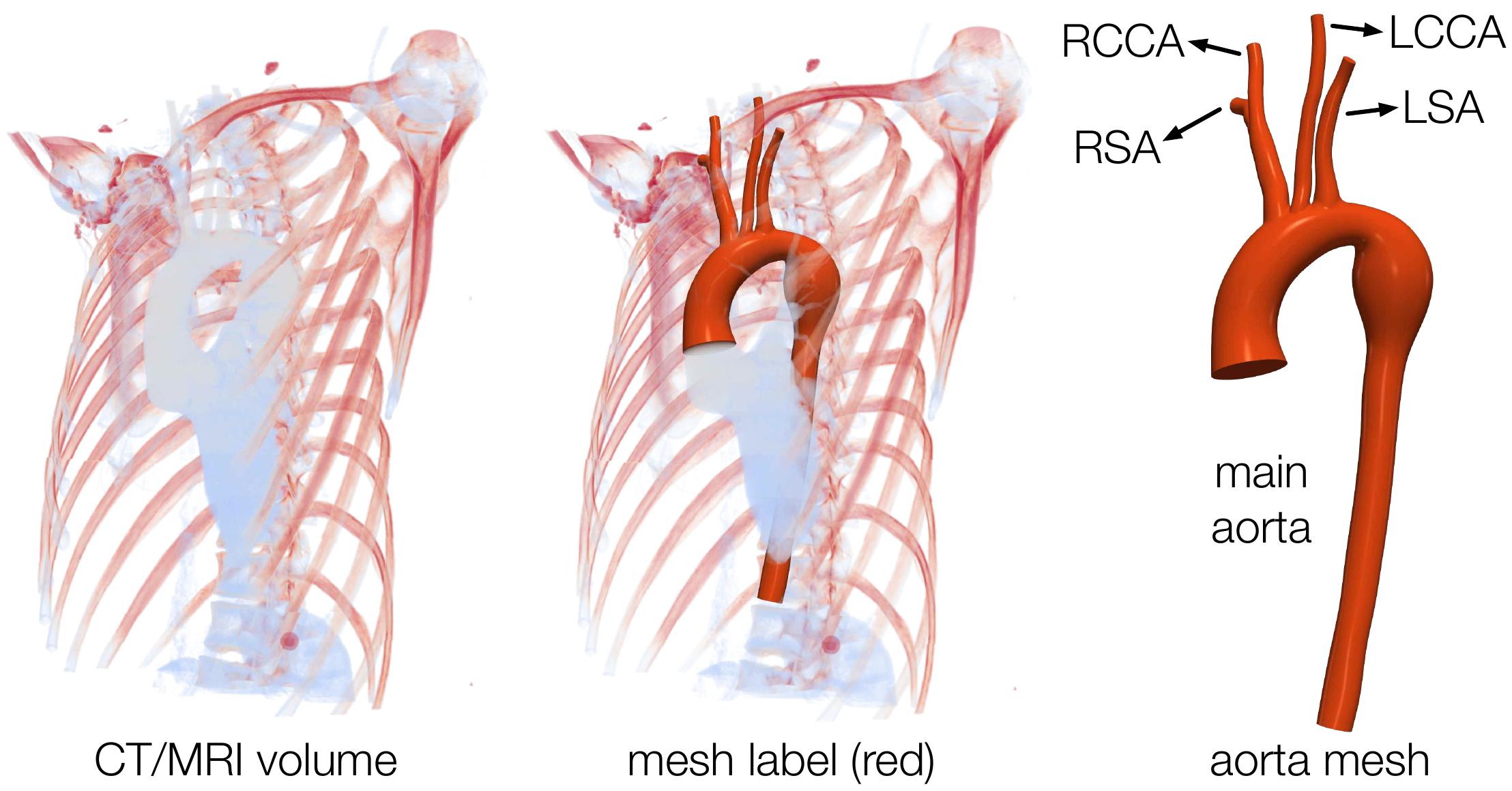}
    \vspace{-0.1in}
    \caption{Extraction of 3D aorta meshes from a CT/MRI volume.}
    \label{fig:aorta}
\end{figure}

The aorta is the largest blood vessel in the human body, originating from the heart and branching out to deliver oxygenated blood throughout the body. 
\hot{Like other blood vessels, it is susceptible to various pathological conditions, such as coarctation and aneurysm~\cite{Aneurysm}. 
Coarctation arises from congenital defects causing aortic narrowing, while aneurysm results from chronic stress or atherosclerosis, leading to localized dilation and wall weakening.}
\pin{Anatomically, the aorta exhibits a multi-branch topology with complex surface features, encompassing significant variations in shape and scale. Although aortic modeling has been extensively studied, the characterization and analysis of multiscale aortic structures—particularly those with branching (e.g., the supra-aortic branches at the arch)—remains a nontrivial challenge~\cite{JinAorta,moccia2018blood}.}
In the context of human anatomy, particularly the aortic arch, the major supra-aortic branches include the {\em right subclavian artery} (RSA), {\em left subclavian artery} (LSA), {\em right common carotid artery} (RCCA), and {\em left common carotid artery} (LCCA). 
As illustrated in Figure~\ref{fig:aorta}, these vessels emerge from the aortic arch and supply blood to the head and upper extremities. 
Given the aorta's vital role, hemodynamic analysis is essential for clinical diagnosis and treatment planning to assess blood flow dynamics and identify potential abnormalities.

{\em Computational fluid dynamics} (CFD) simulations are a powerful tool for understanding aortic disease development and rupture risk. 
By computing critical hemodynamic parameters such as blood flow velocity, pressure distribution, and {\em wall shear stress} (WSS), CFD analysis provides valuable insights for medical professionals. 
However, the accuracy and reliability of CFD simulations are highly dependent on the quality of the underlying 3D aorta meshes. These meshes are typically manually extracted from {\em computed tomography} (CT) or {\em magnetic resonance imaging} (MRI) volumes using tools like SimVascular~\cite{updegrove2017simvascular}. 
This manual process is labor-intensive and time-consuming, requiring the annotation of contours across hundreds or even thousands of CT/MRI slices, making it a significant bottleneck in clinical workflows.

\vspace{-0.1in}

To minimize manual intervention, researchers have developed various segmentation methods~\cite{Marija2020, fantazzini20203d, ZHANG2025106991}.
While these methods, typically trained on large annotated CT/MRI datasets, achieve high segmentation accuracy, they fall short in directly constructing 3D aorta meshes. 
Point clouds offer a more accurate approximation of 3D aortic structures than semantic segmentation. 
Advances in computer vision have led to point cloud-based methods~\cite{LuoH21, PVD} that perform well in generating general 3D objects, such as chairs, aircraft, and cars. 
However, these methods rely on large-scale training datasets, which are significantly harder to obtain for aortic data. 
A more direct approach is to generate 3D surfaces. 
Studies~\cite{AuerG10, Ye_2023_ICCV} have shown that using statistical shapes and deformable models can effectively construct specific organs, such as the heart, due to their relatively simple structure. 
\pin{For modeling complex vascular structures, recent studies~\cite{BloodVesselGAN, DanuNVSI19, KuipersKMB24} have combined generative models (e.g., VAEs, GANs, diffusion models) with statistical shape modeling to synthesize multi-branch vessel geometries. However, these methods primarily focus on small-scale vasculatures and branching topology, often lacking detailed surface representations. Moreover, they serve purely as geometry synthesizers without leveraging medical imaging data.}
\hot{Mistelbauer et al.\ \cite{EFDs} employed {\em elliptic Fourier descriptors} (EFDs) to generate surface meshes for aortic dissections with separated lumens. 
However, creating CFD-ready meshes for healthy aortas and other pathologies—featuring multiple branches and varying diameters—remains a significant challenge, demanding more flexible and generalizable modeling techniques.}


In summary, existing solutions face three major challenges in fully automating aorta mesh extraction from CT/MRI volumes:
\begin{myitemize}
\vspace{-0.05in}
\item {\bf Limited availability of aortic data}:\ Publicly accessible datasets primarily provide segmentation labels rather than detailed 3D mesh annotations, restricting the development of fully automated construction methods.
\item {\bf High reliance on manual intervention}:\ Generating high-quality aorta meshes still requires considerable manual effort, \pin{including mesh smoothing, hole filling, and geometric refinement,} making the process labor-intensive and time-consuming.
\item {\bf Lack of an end-to-end solution}:\ \pin{Existing methods cannot directly generate meshes from medical images that are suitable for both visualization and hemodynamic simulation of the main aorta and supra-aortic branches, necessitating post-processing to obtain a complete and usable 3D representation.}
\vspace{-0.05in}
\end{myitemize}

To address these challenges, we propose AortaDiff, an automatic framework for aorta mesh construction conditioned on CT/MRI volumes. 
AortaDiff integrates both point cloud and segmentation-based approaches and consists of three key stages: 
(1) A volume-guided {\em conditional diffusion model} (CDM) generates the aortic centerline from the input CT/MRI volume.
(2) A contour extraction technique uses the generated centerline points as priors to guide a semi-automated segmentation method, transforming it into a fully automated process.
(3) \pin{A surface construction method constructs the final aorta mesh by fitting a 3D {\em non-uniform rational B-spline} (NURBS) surface to the extracted contours.}
These stages are designed to \pin{address} the limitations of existing approaches.
In the first stage, inspired by the success of {\em denoising diffusion probabilistic models} (DDPM)~\cite{DDPM}, we employ a CDM~\cite{CDM} to generate centerline points rather than the entire \pin{aortic} point cloud. 
This approach provides a compact yet morphologically informative representation of the aorta, significantly reducing computational overhead and enabling effective training on limited aortic datasets. 
In the second stage, the generated centerline points serve as prompts for a prompt-dependent semi-automated segmentation model, eliminating manual intervention in the segmentation process. 
The third stage completes the construction by generating a \pin{CFD-ready} aorta mesh, making AortaDiff a fully end-to-end solution for aorta mesh construction.

The primary contributions of AortaDiff are as follows:
\begin{myitemize}
\vspace{-0.025in}
\item \textbf{Fully automatic aorta mesh extraction}:\ AortaDiff offers an end-to-end solution for generating 3D aorta meshes directly from CT/MRI volumes, covering \pin{normal, aneurysmal, and coarctation cases, and removing} the need for extensive manual intervention.
\item \textbf{Minimal training data requirements}:\ Unlike deep learning-based segmentation and point cloud generation methods, AortaDiff operates effectively without requiring large annotated training datasets, making it practical for real-world applications.
\item \textbf{High-quality mesh generation}:\ \delin{Parameterized NURBS-based meshes produced by AortaDiff are directly compatible with CFD solvers without additional post-processing, enabling efficient hemodynamic analysis, including the estimation of flow-derived biomarkers critical for diagnosis and treatment planning.} 
\item \delin{\textbf{Support for parametric anatomical modification}:\ The flexible geometry construction framework enables customizable anatomical adjustments, facilitating in silico experimentation for device optimization and surgical planning.}
\vspace{-0.05in}
\end{myitemize}
\noindent \hot{By addressing key limitations of existing methods, AortaDiff provides a structured, fully automated approach to aorta mesh construction that renders manual contour annotation and post-processing unnecessary, thereby streamlining the workflow and enhancing the efficiency and consistency of hemodynamic analysis in clinical applications.}

\vspace{-0.075in}
\section{Background and Related Work}

\textbf{Background.} 
\pin{Hemodynamic analysis is essential for understanding cardiovascular conditions and guiding diagnosis and treatment decisions. It enables the estimation of biomarkers and flow characteristics linked to disease progression, such as regions of elevated WSS associated with thrombus formation and aneurysm rupture risk, and abnormal pressure gradients or flow acceleration that help assess aortic coarctation severity and plan interventions. These analyses, typically based on CFD simulations of cardiovascular geometries, provide quantitative insights into vessel morphology, flow velocity, pressure, and WSS distributions, offering biomarkers beyond vessel diameter to better predict complications and evaluate the impact of surgical or endovascular treatments. Parametric geometry models further support in silico experimentation for optimizing device designs and planning procedures.}

Numerous studies have explored visualizing and analyzing hemodynamic data for clinical applications. 
Lawonn et al.\ \cite{LawonnGVP016} developed a tool for occlusion-free blood flow animation to enhance visualization clarity. 
Oeltze-Jafra et al.\ \cite{Oeltze-JafraCJP16} introduced a clustering-based approach for visualizing vortical flows in CFD simulations. 
Tao et al.\ \cite{TaoHQWJSZY16} designed a web-based interface to visualize 4D hemodynamic data.
Combining patient-specific hemodynamics with wall deformation, Meuschke et al.\ \cite{MeuschkeVBPL17} developed a visualization tool integrating 2.5D and 3D representations. 
To support the exploration of multi-field aneurysm data, they also proposed an interactive tool to identify regions of interest on vessel walls~\cite{MeuschkeGBWPL19}. 
Other notable contributions include CAVOCLA~\cite{MeuschkeOBPL19}, which facilitates blood flow classification for cerebral aneurysms, and GUCCI~\cite{GUCCI}, which identifies cohort-specific characteristics of the aorta using glyph-based depictions.

Despite these advancements, a significant gap remains in obtaining high-quality vascular geometries, which are fundamental for CFD simulations~\cite{du2025ai, du2025hugvas}. 
The aorta, the largest blood vessel originating from the heart, has a complex structure, including the main and supra-aortic branches. 
Accurate 3D representations of aortic geometries are critical for clinical applications. 
However, constructing these geometries poses several challenges, including the complexity of multi-branch structures, the limited availability of medical imaging data, and the demand for high geometric fidelity.

\textbf{Surface construction.} 
Recent advancements in computer vision and deep learning have demonstrated significant potential in medical image processing and 3D surface construction. 
Most approaches primarily concentrate on image or volume segmentation to extract 3D surfaces, while others focus on the 3D construction of specific organs, such as the heart and liver. 
More recently, diffusion models have been utilized to generate 3D point clouds~\cite{LuoH21, Wu23}. 
Existing methods for 3D surface construction can be broadly classified into three categories: {\em segmentation}, {\em point cloud}, and {\em geometry} techniques.

\textbf{Segmentation methods} are used to identify vascular regions in 3D medical images, generating segmented volumes of specific organs. 
Techniques such as marching cubes~\cite{Marchingcubes} \pin{or surface nets~\cite{SurfaceNet}}, are then employed to extract 3D surfaces. 
Advanced 3D segmentation models, like DenseUNet~\cite{DenseUNet}, utilize densely connected layers to improve feature propagation, particularly for complex structures. 
UNetR++~\cite{UNetR++}, which incorporates transformer architectures to capture long-range dependencies, achieves state-of-the-art performance. 
Numerous studies have applied segmentation to vascular analysis. 
For instance, Lin et al.\ \cite{LinLGG22} introduced a geometry-constrained deformable attention network to enhance aortic structure segmentation. 
Montalt-Tordera et al.\ \cite{Montalt2022} trained a U-Net model for thoracic aorta hemodynamics analysis, tackling patient-specific blood flow assessment challenges. 
Myronenko et al.\ \cite{MyronenkoYHX23} developed an automated method for aorta segmentation using 3D CT images. 
Vagenas et al.\ \cite{VagenasGM23} combined traditional image processing and machine learning techniques for precise aorta segmentation and construction. 
While these methods effectively capture detailed anatomical features, they rely on extensive annotated data and are susceptible to error propagation, limiting their utility in aortic surface construction. 
\hot{Although region-growing~\cite{region-growing} methods have advanced aortic segmentation, they require seed point specification and are highly sensitive to intensity thresholds, affecting accuracy and consistency. 
{\em Segment anything model} (SAM)-based methods~\cite{SAM} rely on precise input prompts to achieve accurate segmentation, still requiring substantial human intervention.}

\begin{figure*}[htbp]
    \centering
    \includegraphics[width=0.85\linewidth]{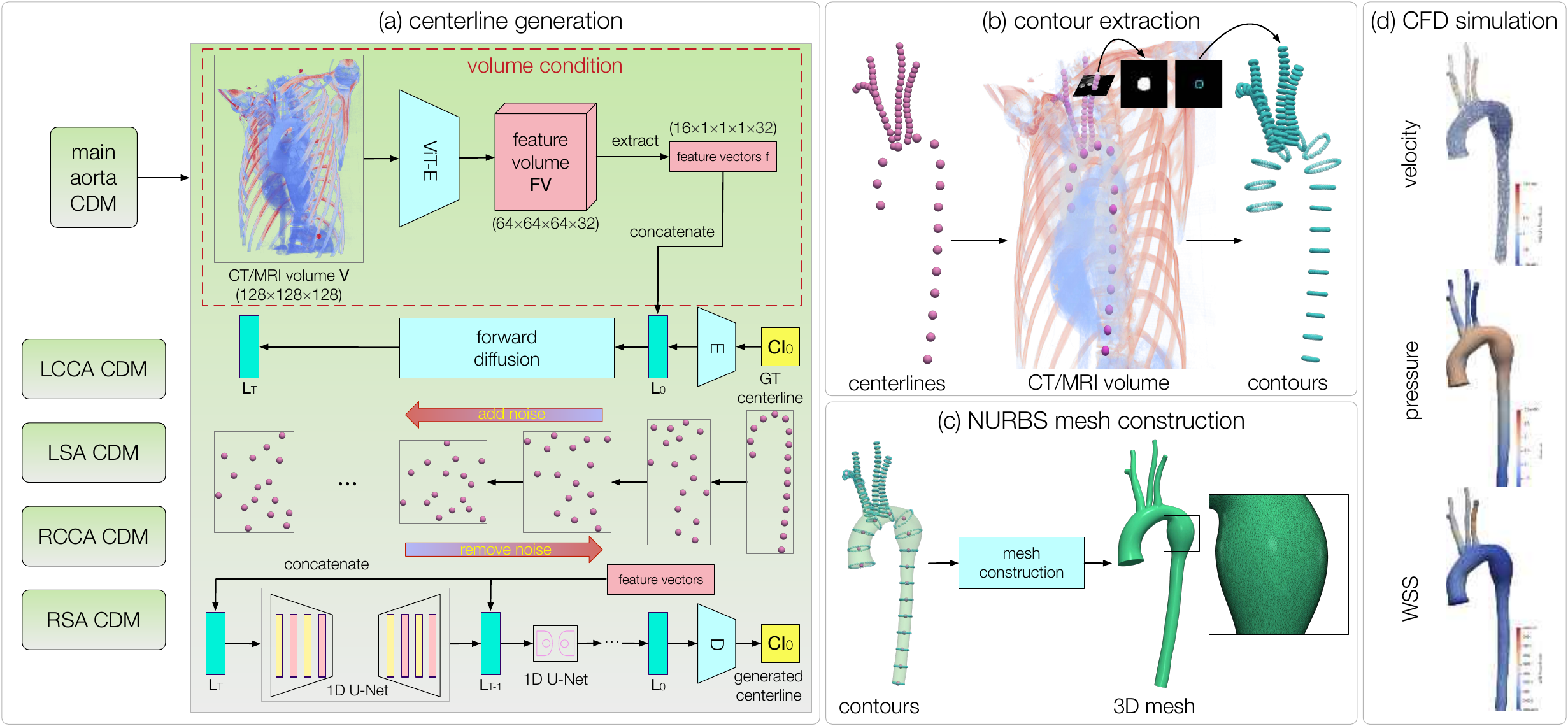} 
    \vspace{-0.1in}
    \caption{The framework of AortaDiff: (a) A volume-guided CDM generates a centerline by extracting a feature volume using a ViT encoder, retrieving feature vectors at the centerline points, and concatenating them with the 1D centerline image. (b) ScribblePrompt extracts contours for centerline points. (c) A 3D NURBS mesh is constructed from the contours representing the geometry. (d) The aorta mesh is then used in CFD simulations.}
    \label{framework}
\end{figure*}

\textbf{Point cloud methods} construct 3D objects by approximating surfaces using sampled points. 
Generative models, particularly diffusion-based approaches, enhance the fidelity of these constructed geometries. 
For example, Erler et al.\ \cite{Erler2020} developed Points2Surf, which learns implicit surface representations from point cloud patches for high-quality construction. 
Zhou et al.\ \cite{PVD} introduced the {\em point-voxel diffusion} (PVD) model, combining denoising diffusion models with point-voxel representations to achieve high-fidelity construction. 
Luo and Hu~\cite{LuoH21} approached point cloud generation as a thermodynamic process, improving robustness against noise. 
Li et al.\ \cite{LiLHF21} proposed a disentangled refinement method for point cloud upsampling, resulting in higher resolution and accuracy. 
Wu et al.\ \cite{Wu23} introduced {\em point straight flow} (PSF), simplifying the diffusion process into a single step, significantly accelerating generation while preserving quality. 
Despite their strengths, these methods require extensive training data and face challenges to ensure continuity and smoothness in constructed surfaces, particularly for complex geometries like the \pin{main aorta and its supra-aortic branches.}

\textbf{Geometry methods} bypass intermediate steps like segmentation or point cloud creation, directly generating 3D geometries. 
For example, \hot{Selle et al.\ \cite{liverVessels} and Hahn et al.\ \cite{HahnPSP01} proposed pipelines for vascular analysis and visualization that combine region-growing-based segmentation with skeletonization and graph-based structural analysis. 
These methods produce effective visualizations of vessel branching patterns and diameters; however, their performance is limited by factors such as low image resolution, partial volume effects, and discontinuities in vessel diameter estimation.} 
Kretschmer et al.\ \cite{KretschmerGPS13} generated and refined smooth vascular models using centerline-based descriptions. 
Recent advancements in deep learning have further enhanced the capabilities of geometry methods. 
Lyu et al.\ \cite{lyu2023} utilized GAN-based models to generate geometries for the aorta and carotid arteries. 
Xiong et al.\ \cite{xiong2022} leveraged cascaded GANs for high-fidelity aortic geometry construction. 
Black et al.\ \cite{black2023} constructed arterial geometries using 4D flow-MRI data to validate hemodynamic simulations. 
Although these methods offer end-to-end solutions, they often require extensive datasets, rely on structural assumptions, and lack flexibility for adjustments.

To address the challenges of aortic surface construction, we propose AortaDiff, a volume-guided CDM that decomposes the aorta into centerlines and contours for precise modeling. 
Diffusion models have shown significant success in image generation, with DDPM~\cite{DDPM} serving as a foundational framework. 
Nichol and Dhariwal~\cite{NicholD21} improved DDPM by modeling reverse diffusion variances, while Chen et al.\ \cite{Image_Diff} introduced dynamic resolution image generation using neural fields. 
Beyond images, diffusion models have been applied to text-to-image generation~\cite{Text-to-Image} and video synthesis~\cite{BIVDiff}. 
In visualization, tools like GAN Lab~\cite{GANLab} have been used to interpret deep generative models. 
\delin{While diffusion models have shown promise in image generation, their application to vessel modeling and visualization remains underexplored. Traditional aorta modeling methods, such as voxel-based reconstruction or manual parameterization, often introduce surface defects or require intensive manual effort, hindering scalability. AortaDiff integrates a volume-guided CDM for centerline generation with SAM-based contour extraction. By combining diffusion modeling with automated parameterization, it generates smooth, simulation-ready meshes with minimal manual intervention, enabling scalable geometry construction for CFD and surgical planning.}

\vspace{-0.1in}
\section{AortaDiff}
\vspace{-0.025in}

Figure~\ref{framework} illustrates the AortaDiff framework, which consists of four key stages:
(a) extracting the aortic centerline from CT/MRI volumes using a CDM,
(b) obtaining vessel contours from cross-sections through image segmentation and edge detection,
(c) constructing the 3D aorta mesh using NURBS, and
(d) conducting downstream CFD simulation tasks with the generated aorta meshes.
\hot{The algorithm description is provided in Section 1 of the appendix.}

\vspace{-0.075in}
\subsection{Centerline Representation}
\vspace{-0.025in}
\label{sec:cl_rep}

Point cloud methods are commonly used for 3D object construction but typically require large training datasets, often numbering in the thousands.
In contrast, aortic datasets are generally limited (e.g., only tens of samples in our experiment), making direct learning from aortic point clouds challenging.
To address this, we adopt a simple yet effective representation: {\em centerlines} and their corresponding {\em contours}.
The centerline represents the central trajectory of the aorta as a 1D approximation of its 3D structure.
It consists of a sequence of ordered points, each corresponding to a specific location along the vessel.
This representation captures the global shape and topology of the aorta while significantly reducing data complexity.
Compared to point clouds, the centerline is more compact, easier to process, and requires fewer computational resources.
Additionally, it integrates seamlessly with other geometric features, such as cross-sectional contours, offering a more efficient and informative vessel representation.
Empirically, we select 16 points per centerline for both the main aorta and supra-aortic branches, striking a balance between accuracy and efficiency.
For the main aorta, this configuration effectively captures its curvature and branching patterns while avoiding unnecessary complexity or overlapping contours that could arise from overly dense sampling.
For the supra-aortic branches, although they are shorter in length, they exhibit greater relative radius variation compared to the main aorta.
\hot{Using 16 points ensures that sufficient detail is preserved while preventing excessive redundancy, which is supported by the study shown in Section 4 of the appendix.}
Unlike point clouds, we encode the centerline as a 1D image—a matrix with three channels corresponding to the centerline points' $x$, $y$, and $z$ coordinates.
This format aligns well with image processing techniques and deep learning models, such as 1D-UNet, a variant of the U-Net architecture~\cite{U-Net}, for centerline generation.

\vspace{-0.05in}
\subsection{Volume-Guided CDM}

To generate the centerline, we employ a volume-guided CDM that progressively denoises a sequence of noisy centerline images.
The CDM is trained to estimate the noise added at each timestep of the diffusion process, enabling it to learn the underlying aortic structure.
Diffusion models, such as DDPM~\cite{DDPM}, are generative frameworks that iteratively reverse a noise diffusion process.
During training, noise is incrementally added to the image, and the model is optimized to predict and remove this noise at each timestep.
This process allows the generation of new samples by starting from a simple prior distribution (e.g., Gaussian) and iteratively refining them through denoising.

However, clinical applications often require constructing a patient-specific 3D aortic surface.
We incorporate information from the 3D CT/MRI volume into the diffusion process to ensure the generated centerline aligns with the underlying anatomical structure.
As shown in Figure~\ref{framework} (a), a {\em vision transformer} (ViT)~\cite{ViT} encoder extracts a feature volume from the input CT/MRI volume, encoding high-level, voxel-wise semantic information.
\hot{This ViT, adapted from MONAI for 3D medical imaging, consists of 12 layers, 8 attention heads, a hidden size of 16, and a multi-layer perceptron dimension of 3072.}
We aim to integrate this feature information for the centerline points as conditional inputs to guide the diffusion process.
A na{\"i}ve approach would associate the entire feature volume with each centerline point.
In contrast, for efficiency and effectiveness, we look up the corresponding feature vector for each centerline point from the feature volume and concatenate it with the noisy centerline image's latent vector.

    {\bf CDM training.}
AortaDiff integrates volumetric features with centerline points at each timestep of the diffusion process:
\begin{myitemize}
    \vspace{-0.05in}
    \item {\bf Input representation}:\ The 1D {\em ground-truth} (GT) centerline image, $\mathbf{CI}_0$, is encoded into a latent vector, $\mathbf{L}_0$. Noise is progressively added to $\mathbf{L}_t$ at each timestep $t$, producing a noisy sequence $(\mathbf{L}_0, \ldots, \mathbf{L}_t, \ldots, \mathbf{L}_T)$. The model input consists of $\mathbf{L}_t$ concatenated with feature vectors, $\mathbf{f}$, extracted from the feature volume, $\mathbf{FV}$.
    \item {\bf Noise prediction}:\ The model is trained to predict the noise, $\epsilon_\theta(\mathbf{CI}_t, t, \mathbf{f})$, added at $t$, facilitating the denoising process.
    \item {\bf Optimization objective}:\ The training aims to minimize the difference between the predicted and actual noise using the loss
    \vspace{-0.05in}
    \begin{equation}
        \mathcal{L}_{\text{CDM}} = \mathbb{E}_{\mathbf{CI}_0, \epsilon, t} \left[ \|\epsilon - \epsilon_\theta(\mathbf{CI}_t, t, \mathbf{f})\|^2 \right],
        \label{eqn:loss}
    \end{equation}
    where $\mathbf{CI}_0$ and $\mathbf{CI}t$ denote the clean and noisy centerline images, $\epsilon$ is the added noise, and $\epsilon\theta$ is the noise predicted by the model.
    \vspace{-0.05in}
\end{myitemize}

{\bf Inference with CDM.}
During inference, the input consists of $\mathbf{L}_T$ (the fully noisy latent vector) and the feature volume, $\mathbf{FV}$.
The process follows the reverse of the training paradigm: at each timestep $t$, the noisy latent vector, $\mathbf{L}_t$, determines voxel locations, and relevant feature vectors, $\mathbf{f}$, are retrieved as conditional inputs.
The model then iteratively denoises the latent vector.
Eventually, at timestep 0, the final output, $\mathbf{CI}_0$, is constructed via a decoder.

\vspace{-0.05in}
\subsection{Contour Extraction}

Extracting contours around centerline points enables the construction of a 3D surface that accurately captures the detailed geometry of the vessel lumen.
We utilize ScribblePrompt~\cite{ScribblePrompt}, a segmentation model that requires minimal user input (e.g., scribbles, clicks, and bounding boxes) to achieve precise segmentation.

    {\bf Orthogonal slice extraction.}
For each centerline point $p$, we compute its tangent vector $\mathbf{t}_p$ using adjacent points.
To maintain computational consistency, the last centerline point shares the tangent vector of its preceding one.
Based on $\mathbf{t}_p$, we define a slicing plane $\Pi_p$ orthogonal to $\mathbf{t}_p$.
The plane's orientation is determined by a {\em local coordinate frame} $(\mathbf{t}_p, \mathbf{n}_p, \mathbf{b}_p)$, where $\mathbf{n}_p$ and $\mathbf{b}_p$ are the normal and binormal vectors spanning $\Pi_p$.
We construct the rotation matrix $\mathbf{R}_p = \begin{bmatrix} \mathbf{b}_p & \mathbf{n}_p & \mathbf{t}_p \end{bmatrix}$, which aligns $\Pi_p$ with the {\em global coordinate system}.
To extract a cross-sectional image $\mathbf{S}_p$ containing the vessel, 
we resample the input volume $\mathbf{V}$ onto $\Pi_p$ using trilinear interpolation
\vspace{-0.05in}
\begin{equation}
    \mathbf{S}_p = \mathbf{V}(\mathbf{R}_p \cdot \mathbf{l}_q + \mathbf{g}_p),
    \label{eqn:sp}
\end{equation}
where $\mathbf{g}_p$ and $\mathbf{l}_q$ denote the coordinates of $p$ in the global coordinate system and any point $q$ on $\Pi_p$ in the local coordinate frame.

\begin{figure}[htb]
    \centering
    \includegraphics[width=0.85\linewidth]{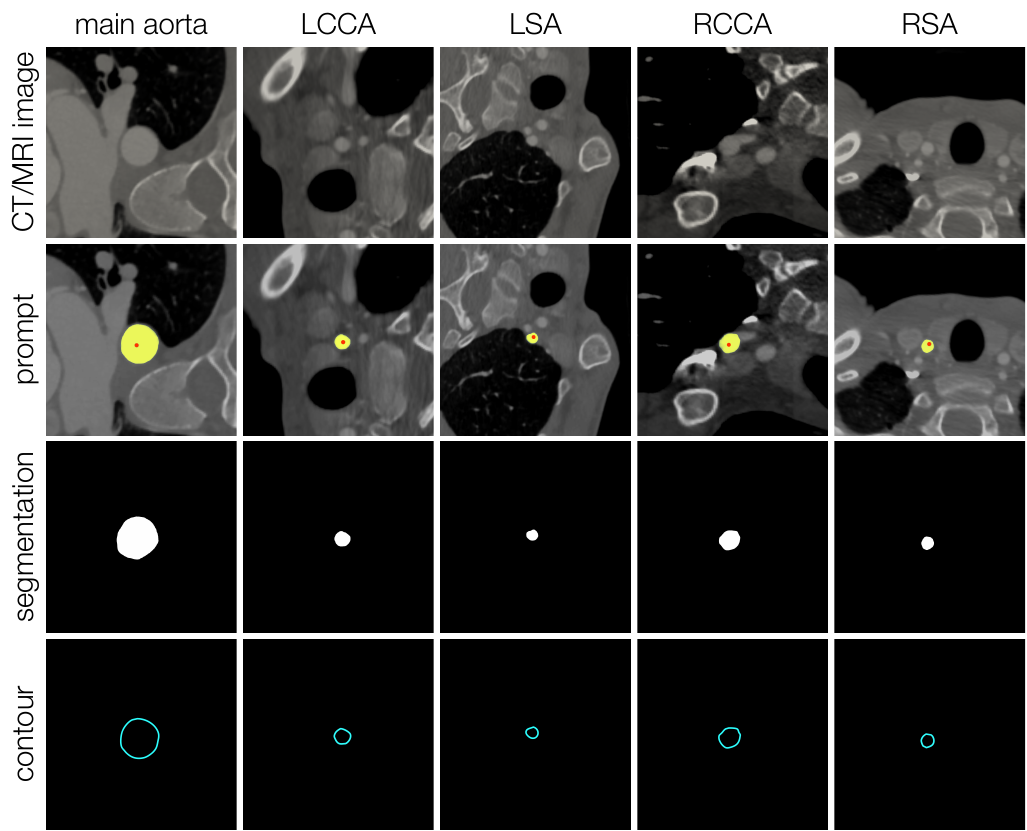} 
    \vspace{-0.1in}
    \caption{Examples of segmentation results using ScribblePrompt. \hot{The red point in the second row denotes the centerline point used as the prompt.}}
    \label{contour}
\end{figure}

    {\bf Segmentation with ScribblePrompt.}
\hot{The extracted slice $\mathbf{S}_p$ is segmented using ScribblePrompt, a SAM-based model trained on large-scale medical datasets with simulated scribbles, clicks, and bounding boxes to mimic realistic user interactions.
We chose ScribblePrompt for its high segmentation accuracy and robustness to slight variations in input prompts.}  
While ScribblePrompt typically relies on \pin{scribble-based input,} we automate this step by directly using the centerline points generated from AortaDiff as prompts.
This results in precise lumen segmentation, effectively adapting to anatomical variations, including geometric transitions like bifurcation.
Examples of the segmentation results are shown in Figure~\ref{contour}.

\vspace{-0.1in}

{\bf Contour extraction and processing.}
\hot{After obtaining the segmentation mask, we extract strong gradients corresponding to the vessel boundary by Canny edge detection.}
The detected edges are then processed with OpenCV's \texttt{findContours} function~\cite{OpenCV}, which retrieves the outermost contour $\mathcal{C}^{\twod}_p$ in a topologically consistent manner.
\delin{Since NURBS requires a consistent number of control points across cross-sections, we uniformly resample $\mathcal{C}^{\twod}_p$ to 32 points. This conservative choice exceeds the typical range used in manual segmentation (5 to 20 points), ensuring sufficient geometric accuracy for reliable hemodynamic analysis. It also maintains smooth, watertight surfaces, even when handling low-resolution or imperfect inputs.} 

    {\bf Mapping contour back to 3D.}
Each 2D contour $\mathcal{C}^{\twod}_p$ extracted from $\Pi_p$ must be mapped back to 3D.
Since $\Pi_p$ is defined by the local coordinate frame $(\mathbf{t}_p, \mathbf{n}_p, \mathbf{b}_p)$ and centered at $p$, we transform each 2D contour point $\mathcal{C}^{\twod}_{p_i}$ back to $\mathcal{C}^{\threed}_{p_i}$ in the original 3D space using the inverse of $\mathbf{R}_p$, i.e., $\mathbf{R}_p^{-1} \cdot \mathbf{l}_c + \mathbf{g}_p$,
where $\mathbf{l}_c$ denotes the coordinate of any contour point $c$ in the local coordinate frame.
This process ensures that the constructed contours align accurately with the input volume $\mathbf{V}$, preserving the anatomical vessel structure while maintaining a consistent contour representation for NURBS-based construction.

\vspace{-0.075in}
\subsection{Aortic Surface Construction}
\vspace{-0.025in}

After obtaining the aortic centerline and corresponding contours, we construct the 3D aortic surface using NURBS, achieving a smooth and anatomically accurate representation of the vessel. 

    {\bf Contour alignment.}
To ensure consistency across adjacent contours, we establish point-to-point correspondence using the {\em iterative closest point} (ICP) algorithm~\cite{ICP}, which preserves spatial coherence along the vessel and provides a reliable foundation for surface construction.

    {\bf Surface fitting.}
We implement a curve-fitting algorithm~\cite{piegl1996nurbs} to fit a cubic spline curve from the aligned contour points $\{\mathcal{C}^{\threed}_{p_i}\}$, which gives the control points $\mathbf{p}_{i,j}\in\mathcal{R}^{m\times n \times 3}$ and weights $w_{i,j} \in \mathcal{R}^{m\times n}$
    \vspace{-0.05in}
    \begin{equation}
        w_{i,j}, \mathbf{p}_{i,j} = \cf\left(\{\mathcal{C}^{\threed}_{p_i}\}\right),
        \label{eqn:sf-1}
    \end{equation}
    where $m$ and $n$ represent the resolutions along the longitudinal (streamwise) and circumferential directions, respectively.
    $i$ denotes the longitudinal position along the centerline, while $j$ represents the circumferential position around each cross-section.
    The estimated control points and weights can be directly used to create a 3D NURBS surface    
    \vspace{-0.05in}
\begin{equation}
    \mathbf{S}(u, v) = \frac{\sum_{i=0}^{n}\sum_{j=0}^{m} N_{i,p}(u) N_{j,q}(v) w_{i,j} \mathbf{p}_{i,j}}{\sum_{i=0}^{n}\sum_{j=0}^{m} N_{i,p}(u) N_{j,q}(v) w_{i,j}},
    \label{eqn:sf-2}
\end{equation}
where $u$ represents the parametric direction along the aortic centerline, capturing the longitudinal dimension, and $v$ represents the circumferential direction around each cross-sectional contour.
$N_{i,p}(u)$ and $N_{j,q}(v)$ are B-spline basis functions, while $w_{i,j}$ and $\mathbf{p}_{i,j}$ represent the corresponding weights and control points. $p$ and $q$ denote the spline orders on $u$ and $v$ directions.
The surface fitting process interpolates control points along $u$ (centerline) and $v$ (contours), generating a continuous parametric surface. This ensures the constructed aorta maintains geometric accuracy, effectively capturing subtle variations in vessel curvature, diameter, and morphology.

    {\bf Multi-branch integration.}
We separately model the main aorta and supra-aortic branches and seamlessly integrate them into a unified surface.
Overlapping regions are refined to eliminate discontinuities. 
NURBS provides a robust framework for vascular modeling, capable of capturing complex branching structures and detailed geometric variations with high precision.

\vspace{-0.05in}
\subsection{CFD Simulation}

To validate the feasibility of the constructed aorta meshes for downstream CFD simulations, we employ OpenFOAM~\cite{OpenFOAM}, a widely used solver for simulating fluid dynamics in complex geometries.
The simulation process consists of the following steps:
\begin{myitemize}
    \vspace{-0.05in}
    \item \textbf{Geometry and mesh generation}: \pin{The constructed aorta geometry is remeshed into a high-quality triangulated surface using PyACVD~\cite{ValetteCP08}, followed by volumetric tetrahedral meshing via VMTK's TetGen-based algorithm~\cite{AntigaPBERS08}. The mesh is compatible with OpenFOAM and dense enough to capture key flow characteristics.} The mesh comprises polyhedral elements with appropriate boundary definitions for the inlet, outlets, and vessel walls.
    \item \textbf{Governing equations and numerical discretization}:\ Each simulation solves steady-state incompressible Navier-Stokes equations
        \vspace{-0.05in}
        \begin{equation}
            \begin{split}
                \nabla\cdot\bm{u} &= 0,\\
                (\bm{u} \cdot \nabla)\bm{u} &= -\nabla p + \nu\nabla^2\bm{u},
                \label{eq:ns}
            \end{split}
        \end{equation}
        where $\mathbf{u}$ is the velocity field, $p$ is the pressure, $\nu$ is the kinematic viscosity. We solve those equations using SimpleFoam on multiple GPUs in parallel via OpenMPI~\cite{graham2006open}. 
    \item \textbf{Boundary conditions}:\ Physiological boundary conditions are applied to replicate realistic aortic flow dynamics, including
        {\em inlet} (a prescribed velocity profile simulating either pulsatile or steady inflow conditions),
    {\em outlets} (either zero-gradient conditions or specified pressure values to allow natural flow development), and
        {\em vessel walls} (a no-slip boundary condition enforcing fluid adherence to the vessel wall).
    \item \textbf{Solver configuration}:\ The simulation employs the {\em semi-implicit method for pressure-linked equations} (SIMPLE) algorithm~\cite{patankar1983calculation} in a steady-state formulation. To enhance computational efficiency, parallel domain decomposition is applied, distributing the computational workload across multiple processors.
    \item \textbf{Post-processing and flow analysis}:\ Once the simulation converges, key hemodynamic parameters, such as velocity magnitude and pressure distribution, are extracted to quantify fluid behavior. WSS is particularly important for assessing the risk of aneurysm rupture, thrombus formation, and disease progression, making it essential for CFD analysis to ensure the clinical relevance of the computational model.
    \vspace{-0.05in}
\end{myitemize}

\vspace{-0.05in}
\section{Results and Discussion}

We conducted experiments on aortic data to evaluate AortaDiff's performance and validate the design of our framework.
Our evaluation compares AortaDiff against state-of-the-art methods for centerline generation and contour extraction.
The resulting aorta meshes are assessed through both quantitative metrics and qualitative analysis.
To further demonstrate the practical applicability of AortaDiff, we performed CFD simulations on the generated meshes and compared the results with those obtained from GT meshes.

\vspace{-0.05in}
\subsection{Experimental Setup}

To train a 3D mesh model of the aorta, the dataset must meet the following criteria: 
(1) It should contain a sufficient number of aorta models with diverse morphologies.
(2) It should include a variety of aortic structures, such as the main aorta and supra-aortic branches.
(3) It should provide mesh labels for the aorta or, at the very least, segmentation labels.
However, datasets containing aorta models are limited. 
Most available datasets provide only segmentation labels for the main aorta, while mesh labels are rare. 

To facilitate the training of our AortaDiff model, we used the {\em vascular model repository} (VMR) dataset~\cite{Wilson2013}, which includes 3D aorta meshes along with corresponding MRI volumes. 
This dataset features various blood vessels, such as the aorta, cerebral arteries, pulmonary arteries, and coronary arteries. 
In total, it contains 275 blood vessel cases, including 91 aortas from both animals and humans. 
Since the dataset includes non-human aortas, we specifically selected 22 human aorta cases for our study, with 18 used for training and four for inference.

To evaluate the generalizability of AortaDiff, we additionally incorporated several cases from the {\em aortic vessel tree} (AVT) dataset~\cite{Radl2022}, which consists of CT volumes. 
This dataset exclusively contains human aorta models, with a total of 56 cases. 
We selected eight cases with diverse aortic morphologies to ensure a more comprehensive evaluation.
Since the AVT dataset does not provide preexisting aorta meshes, we generated them by converting the segmentation labels into surface meshes using the marching cubes algorithm~\cite{Marchingcubes}. 
We then performed manual post-processing to ensure surface smoothness and to eliminate topological artifacts such as holes.

\hot{All CT/MRI volumes were resampled with an isotropic spacing of $0.8 \times 0.8 \times 0.3$ mm, resized to $128 \times 128 \times 128$, and normalized to an intensity range of $[0,1]$ while retaining the full volume.} 
Data augmentation was applied using MONAI's \texttt{RandFlipd} and \texttt{RandRotate90d}~\cite{MONAI}, each with a probability of 0.1.
Model training was performed on a single NVIDIA A40 GPU with 48 GB of memory.
We used the Adam optimizer with hyperparameters $\beta_1 = 0.9$ and $\beta_2 = 0.99$.
The learning rate was initialized at $1 \times 10^{-3}$ with a batch size of 16.
The training process was run for a total of 100,000 iterations.

\hot{In the paper, the tables give quantitative results for all twelve test cases (four from the VMR dataset and eight from the AVT dataset), while the figures showcase qualitative results for six selected cases. The remaining six cases are included in Section 3 of the appendix.}

\begin{figure*}[htb]
    \centering
    \includegraphics[width=1.0\linewidth]{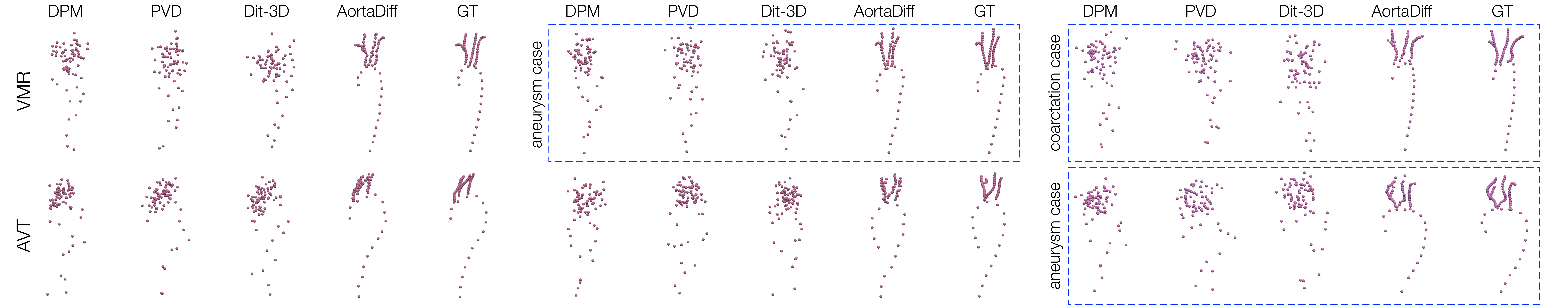} 
    \vspace{-0.25in}
    \caption{\hot{Comparison of centerline generation methods on the selected test cases from the VMR and AVT datasets.}}
    \label{cl-baselines}
\end{figure*}

\vspace{-0.1in}
\begin{table}[htb]
    \caption{\hot{Average CD, HD, and EMD for centerline generation on all test cases from the VMR and AVT datasets. For all tables in this paper, values are in millimeters (mm) as mean ± standard deviation, with the best results highlighted in bold.}}
    \vspace{-0.1in}
    \centering
    \resizebox{\columnwidth}{!}{
        \begin{tabular}{ccccccccc}
            & \multicolumn{4}{c}{\hot{VMR}} & \multicolumn{4}{c}{\hot{AVT}} \\
            \cmidrule(lr){2-5} \cmidrule(lr){6-9}
            \hot{metric} & \hot{DPM} & \hot{PVD} & \hot{Dit-3D} & \hot{AortaDiff} & \hot{DPM} & \hot{PVD} & \hot{Dit-3D} & \hot{AortaDiff} \\ \hline
            & \multicolumn{8}{c}{\hot{supra-aortic branches}} \\
            \hot{CD $\downarrow$}  & \hot{2.68$\pm$0.05} & \hot{2.46$\pm$0.06} & \hot{2.25$\pm$0.09} & \hot{\textbf{0.37$\pm$0.02}} & \hot{3.05$\pm$0.25} & \hot{2.80$\pm$0.13} & \hot{2.33$\pm$0.25} & \hot{\textbf{0.55$\pm$0.02}} \\
            \hot{HD $\downarrow$}  & \hot{3.90$\pm$0.20} & \hot{3.62$\pm$0.15} & \hot{3.12$\pm$0.10} & \hot{\textbf{0.35$\pm$0.01}} & \hot{3.93$\pm$0.15} & \hot{3.87$\pm$0.25} & \hot{4.05$\pm$0.10} & \hot{\textbf{0.37$\pm$0.01}} \\
            \hot{EMD $\downarrow$} & \hot{1.82$\pm$0.05} & \hot{1.75$\pm$0.05} & \hot{1.54$\pm$0.09} & \hot{\textbf{0.19$\pm$0.02}} & \hot{1.57$\pm$0.12} & \hot{1.63$\pm$0.08} & \hot{1.60$\pm$0.15} & \hot{\textbf{0.29$\pm$0.00}} \\
            \hdashline
            & \multicolumn{8}{c}{\hot{main aorta}} \\
            \hot{CD $\downarrow$}  & \hot{2.09$\pm$0.03} & \hot{1.97$\pm$0.05} & \hot{1.71$\pm$0.06} & \hot{\textbf{0.29$\pm$0.01}} & \hot{1.85$\pm$0.20} & \hot{1.88$\pm$0.12} & \hot{1.89$\pm$0.25} & \hot{\textbf{0.44$\pm$0.02}} \\
            \hot{HD $\downarrow$}  & \hot{3.49$\pm$0.15} & \hot{3.11$\pm$0.15} & \hot{2.60$\pm$0.10} & \hot{\textbf{0.26$\pm$0.00}} & \hot{3.30$\pm$0.10} & \hot{3.39$\pm$0.22} & \hot{3.55$\pm$0.10} & \hot{\textbf{0.32$\pm$0.01}} \\
            \hot{EMD $\downarrow$} & \hot{1.40$\pm$0.03} & \hot{1.34$\pm$0.05} & \hot{1.15$\pm$0.05} & \hot{\textbf{0.14$\pm$0.01}} & \hot{1.30$\pm$0.13} & \hot{1.30$\pm$0.05} & \hot{1.30$\pm$0.13} & \hot{\textbf{0.23$\pm$0.02}} \\
            \hdashline
            & \multicolumn{8}{c}{\hot{overall}} \\
            \hot{CD $\downarrow$}  & \hot{2.39$\pm$0.05} & \hot{2.22$\pm$0.06} & \hot{1.98$\pm$0.08} & \hot{\textbf{0.33$\pm$0.02}} & \hot{2.45$\pm$0.23} & \hot{2.34$\pm$0.13} & \hot{2.11$\pm$0.25} & \hot{\textbf{0.50$\pm$0.02}} \\
            \hot{HD $\downarrow$}  & \hot{3.70$\pm$0.18} & \hot{3.37$\pm$0.15} & \hot{2.85$\pm$0.10} & \hot{\textbf{0.31$\pm$0.01}} & \hot{3.62$\pm$0.13} & \hot{3.63$\pm$0.24} & \hot{3.75$\pm$0.10} & \hot{\textbf{0.35$\pm$0.01}} \\
            \hot{EMD $\downarrow$} & \hot{1.60$\pm$0.05} & \hot{1.52$\pm$0.05} & \hot{1.32$\pm$0.07} & \hot{\textbf{0.17$\pm$0.02}} & \hot{1.45$\pm$0.13} & \hot{1.46$\pm$0.06} & \hot{1.45$\pm$0.14} & \hot{\textbf{0.26$\pm$0.03}} \\
        \end{tabular}
    }
    \label{tab:cl-baselines}
\end{table}

\begin{table*}[htbp]
    \caption{\hot{Average Dice, ASD, and HD for segmentation on all test cases from the VMR and AVT datasets.}}
    \vspace{-0.1in}
    \centering
    \resizebox{5in}{!}{
        \begin{tabular}{ccccccccccc}
                             & \multicolumn{5}{c}{\hot{VMR}} & \multicolumn{5}{c}{\hot{AVT}} \\
            \cmidrule(lr){2-6} \cmidrule(lr){7-11}
            \hot{metric} & \hot{SAM} & \hot{MedSAM} & \hot{TotalSeg} & \hot{LoGB-Net} & \hot{AortaDiff} 
                        & \hot{SAM} & \hot{MedSAM} & \hot{TotalSeg} & \hot{LoGB-Net} & \hot{AortaDiff} \\ \hline
            \multicolumn{11}{c}{\hot{supra-aortic branches}} \\
            \hot{Dice $\uparrow$}  & \hot{0.575$\pm$0.078} & \hot{0.757$\pm$0.024} & \hot{0.800$\pm$0.007} & \hot{0.890$\pm$0.006} & \hot{\textbf{0.967}$\pm$\textbf{0.020}} 
                                    & \hot{0.553$\pm$0.075} & \hot{0.736$\pm$0.019} & \hot{0.782$\pm$0.020} & \hot{0.870$\pm$0.018} & \hot{\textbf{0.947}$\pm$\textbf{0.012}} \\
            \hot{ASD $\downarrow$} & \hot{1.720$\pm$0.350} & \hot{0.960$\pm$0.148} & \hot{0.715$\pm$0.160} & \hot{0.705$\pm$0.170} & \hot{\textbf{0.502}$\pm$\textbf{0.035}} 
                                    & \hot{1.638$\pm$0.315} & \hot{0.875$\pm$0.170} & \hot{0.635$\pm$0.150} & \hot{0.620$\pm$0.160} & \hot{\textbf{0.457}$\pm$\textbf{0.055}} \\
            \hot{HD $\downarrow$}  & \hot{9.150$\pm$2.880} & \hot{8.790$\pm$3.000} & \hot{6.580$\pm$0.330} & \hot{6.470$\pm$0.340} & \hot{\textbf{5.752}$\pm$\textbf{0.200}} 
                                    & \hot{8.420$\pm$2.860} & \hot{8.060$\pm$2.990} & \hot{6.100$\pm$0.360} & \hot{6.000$\pm$0.370} & \hot{\textbf{5.312}$\pm$\textbf{0.230}} \\
            \hdashline
            \multicolumn{11}{c}{\hot{main aorta}} \\
            \hot{Dice $\uparrow$}  & \hot{0.605$\pm$0.078} & \hot{0.787$\pm$0.024} & \hot{0.910$\pm$0.007} & \hot{0.918$\pm$0.006} & \hot{\textbf{0.972}$\pm$\textbf{0.020}} 
                                    & \hot{0.565$\pm$0.075} & \hot{0.746$\pm$0.019} & \hot{0.882$\pm$0.020} & \hot{0.880$\pm$0.018} & \hot{\textbf{0.952}$\pm$\textbf{0.012}} \\
            \hot{ASD $\downarrow$} & \hot{1.690$\pm$0.350} & \hot{0.938$\pm$0.148} & \hot{0.705$\pm$0.160} & \hot{0.693$\pm$0.170} & \hot{\textbf{0.490}$\pm$\textbf{0.035}} 
                                    & \hot{1.628$\pm$0.315} & \hot{0.868$\pm$0.170} & \hot{0.625$\pm$0.150} & \hot{0.608$\pm$0.160} & \hot{\textbf{0.450}$\pm$\textbf{0.055}} \\
            \hot{HD $\downarrow$}  & \hot{9.110$\pm$2.880} & \hot{8.740$\pm$3.000} & \hot{6.520$\pm$0.330} & \hot{6.410$\pm$0.340} & \hot{\textbf{5.720}$\pm$\textbf{0.200}} 
                                    & \hot{8.390$\pm$2.860} & \hot{8.030$\pm$2.990} & \hot{6.050$\pm$0.360} & \hot{5.950$\pm$0.370} & \hot{\textbf{5.280}$\pm$\textbf{0.230}} \\
            \hdashline
            \multicolumn{11}{c}{\hot{overall}} \\
            \hot{Dice $\uparrow$}  & \hot{0.590$\pm$0.078} & \hot{0.772$\pm$0.022} & \hot{0.885$\pm$0.007} & \hot{0.904$\pm$0.010} & \hot{\textbf{0.970}$\pm$\textbf{0.016}} 
                                    & \hot{0.559$\pm$0.075} & \hot{0.741$\pm$0.019} & \hot{0.860$\pm$0.020} & \hot{0.880$\pm$0.018} & \hot{\textbf{0.950}$\pm$\textbf{0.012}} \\
            \hot{ASD $\downarrow$} & \hot{1.705$\pm$0.340} & \hot{0.949$\pm$0.160} & \hot{0.710$\pm$0.160} & \hot{0.699$\pm$0.170} & \hot{\textbf{0.496}$\pm$\textbf{0.045}} 
                                    & \hot{1.633$\pm$0.315} & \hot{0.872$\pm$0.170} & \hot{0.630$\pm$0.150} & \hot{0.614$\pm$0.160} & \hot{\textbf{0.454}$\pm$\textbf{0.055}} \\
            \hot{HD $\downarrow$}  & \hot{9.130$\pm$2.880} & \hot{8.765$\pm$3.000} & \hot{6.550$\pm$0.330} & \hot{6.440$\pm$0.340} & \hot{\textbf{5.736}$\pm$\textbf{0.215}} 
                                    & \hot{8.405$\pm$2.860} & \hot{8.045$\pm$2.990} & \hot{6.075$\pm$0.360} & \hot{5.975$\pm$0.370} & \hot{\textbf{5.296}$\pm$\textbf{0.230}} \\
        \end{tabular}
    }
    \label{tab:segmentation-baselines}
\end{table*}

\begin{figure*}[htb]
    \centering
    \includegraphics[width=0.85\linewidth]{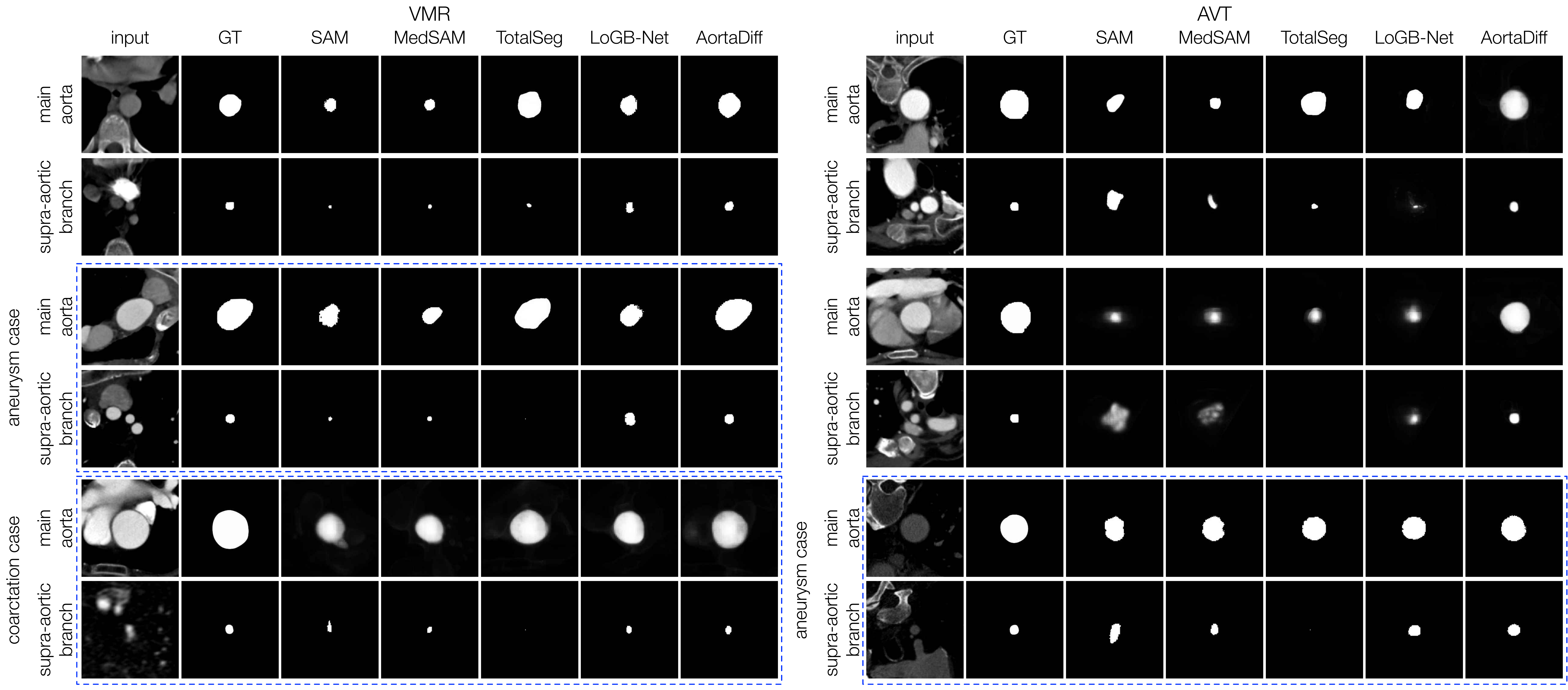} 
    \vspace{-0.1in}
    \caption{\hot{Comparison of segmentation methods on the selected test cases from the VMR and AVT datasets.
        Each case includes segmentation results for both the main aorta and supra-aortic branches.
        Cross-sections are resampled on planes orthogonal to the centerline, with each plane centered on a corresponding centerline point.}}
    \label{segmentation}
\end{figure*}

\vspace{-0.1in}
\begin{figure*}[htb]
    \centering
    \includegraphics[width=1.0\linewidth]{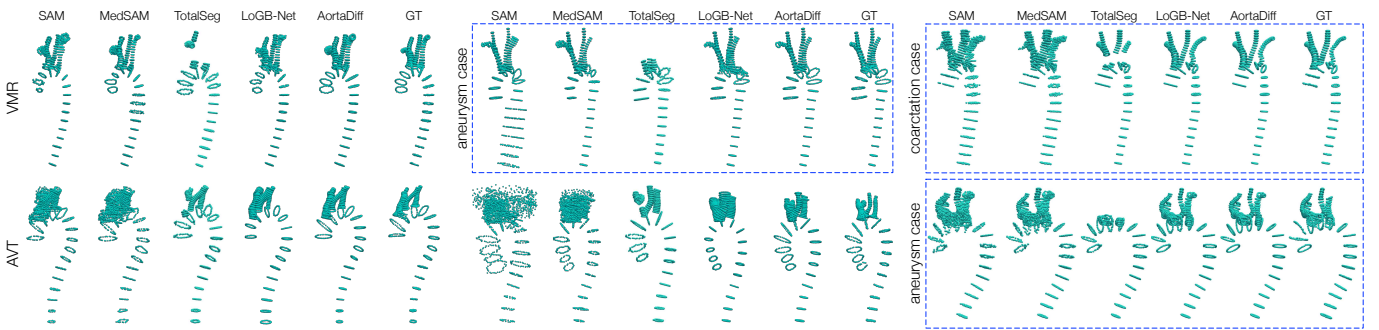} 
    \vspace{-0.25in}
    \caption{\hot{Comparison of contours extracted from different segmentation methods on the selected test cases from the VMR and AVT datasets.}}
    \label{fig:contours-results}
\end{figure*}

\begin{table*}[htbp]
    \caption{\hot{Average CD, HD, and EMD for contour point clouds on all test cases from the VMR and AVT datasets.}}
    \vspace{-0.1in}
    \centering
    \resizebox{4.25in}{!}{
        \begin{tabular}{ccccccccccc}
                             & \multicolumn{5}{c}{\hot{VMR}} & \multicolumn{5}{c}{\hot{AVT}} \\
            \cmidrule(lr){2-6} \cmidrule(lr){7-11}
            \hot{metric} & \hot{SAM} & \hot{MedSAM} & \hot{TotalSeg} & \hot{LoGB-Net} & \hot{AortaDiff}
                         & \hot{SAM} & \hot{MedSAM} & \hot{TotalSeg} & \hot{LoGB-Net} & \hot{AortaDiff} \\ \hline
            \multicolumn{11}{c}{\hot{supra-aortic branches}} \\
            \hot{CD $\downarrow$} & \hot{3.40$\pm$0.55} & \hot{2.85$\pm$0.50} & \hot{3.00$\pm$0.52} & \hot{1.95$\pm$0.35} & \hot{\textbf{1.30}$\pm$\textbf{0.18}}
                                   & \hot{3.65$\pm$0.58} & \hot{3.00$\pm$0.52} & \hot{3.10$\pm$0.55} & \hot{2.10$\pm$0.38} & \hot{\textbf{1.40}$\pm$\textbf{0.20}} \\
            \hot{HD $\downarrow$} & \hot{9.80$\pm$3.10} & \hot{9.10$\pm$3.15} & \hot{9.20$\pm$3.20} & \hot{6.80$\pm$0.55} & \hot{\textbf{6.00}$\pm$\textbf{0.35}}
                                   & \hot{10.20$\pm$3.25} & \hot{9.40$\pm$3.20} & \hot{9.50$\pm$3.30} & \hot{7.10$\pm$0.58} & \hot{\textbf{6.20}$\pm$\textbf{0.38}} \\
            \hot{EMD $\downarrow$} & \hot{1.85$\pm$0.45} & \hot{1.30$\pm$0.35} & \hot{1.35$\pm$0.38} & \hot{1.00$\pm$0.25} & \hot{\textbf{0.68}$\pm$\textbf{0.12}}
                                    & \hot{2.00$\pm$0.48} & \hot{1.40$\pm$0.38} & \hot{1.45$\pm$0.40} & \hot{1.10$\pm$0.28} & \hot{\textbf{0.72}$\pm$\textbf{0.14}} \\
            \hdashline
            \multicolumn{11}{c}{\hot{main aorta}} \\
            \hot{CD $\downarrow$} & \hot{3.20$\pm$0.50} & \hot{2.70$\pm$0.45} & \hot{1.95$\pm$0.32} & \hot{1.85$\pm$0.30} & \hot{\textbf{1.20}$\pm$\textbf{0.12}}
                                   & \hot{3.45$\pm$0.52} & \hot{2.80$\pm$0.48} & \hot{2.10$\pm$0.35} & \hot{1.98$\pm$0.32} & \hot{\textbf{1.30}$\pm$\textbf{0.14}} \\
            \hot{HD $\downarrow$} & \hot{9.60$\pm$3.00} & \hot{8.90$\pm$3.05} & \hot{6.80$\pm$0.52} & \hot{6.60$\pm$0.45} & \hot{\textbf{5.90}$\pm$\textbf{0.27}}
                                   & \hot{10.00$\pm$3.20} & \hot{9.20$\pm$3.15} & \hot{7.20$\pm$0.55} & \hot{6.90$\pm$0.50} & \hot{\textbf{6.05}$\pm$\textbf{0.32}} \\
            \hot{EMD $\downarrow$} & \hot{1.82$\pm$0.40} & \hot{1.28$\pm$0.32} & \hot{1.00$\pm$0.22} & \hot{0.92$\pm$0.19} & \hot{\textbf{0.61}$\pm$\textbf{0.09}}
                                    & \hot{1.95$\pm$0.42} & \hot{1.35$\pm$0.35} & \hot{1.10$\pm$0.24} & \hot{1.02$\pm$0.22} & \hot{\textbf{0.66}$\pm$\textbf{0.11}} \\
            \hdashline
            \multicolumn{11}{c}{\hot{overall}} \\
            \hot{CD $\downarrow$} & \hot{3.28$\pm$0.52} & \hot{2.78$\pm$0.47} & \hot{2.30$\pm$0.40} & \hot{1.90$\pm$0.33} & \hot{\textbf{1.25}$\pm$\textbf{0.15}}
                                   & \hot{3.55$\pm$0.55} & \hot{2.90$\pm$0.50} & \hot{2.45$\pm$0.42} & \hot{2.00$\pm$0.35} & \hot{\textbf{1.35}$\pm$\textbf{0.17}} \\
            \hot{HD $\downarrow$} & \hot{9.70$\pm$3.05} & \hot{9.00$\pm$3.10} & \hot{7.50$\pm$0.55} & \hot{6.70$\pm$0.48} & \hot{\textbf{5.95}$\pm$\textbf{0.30}}
                                   & \hot{10.10$\pm$3.23} & \hot{9.30$\pm$3.18} & \hot{7.40$\pm$0.58} & \hot{7.00$\pm$0.52} & \hot{\textbf{6.10}$\pm$\textbf{0.34}} \\
            \hot{EMD $\downarrow$} & \hot{1.84$\pm$0.42} & \hot{1.29$\pm$0.34} & \hot{1.10$\pm$0.23} & \hot{0.96$\pm$0.20} & \hot{\textbf{0.64}$\pm$\textbf{0.10}}
                                    & \hot{1.98$\pm$0.44} & \hot{1.38$\pm$0.36} & \hot{1.20$\pm$0.25} & \hot{1.05$\pm$0.23} & \hot{\textbf{0.69}$\pm$\textbf{0.12}} \\
        \end{tabular}
    }
    \label{tab:contour_pc}
\end{table*}

\vspace{-0.1in}
\begin{figure*}[htb]
    \centering
    \includegraphics[width=0.85\linewidth]{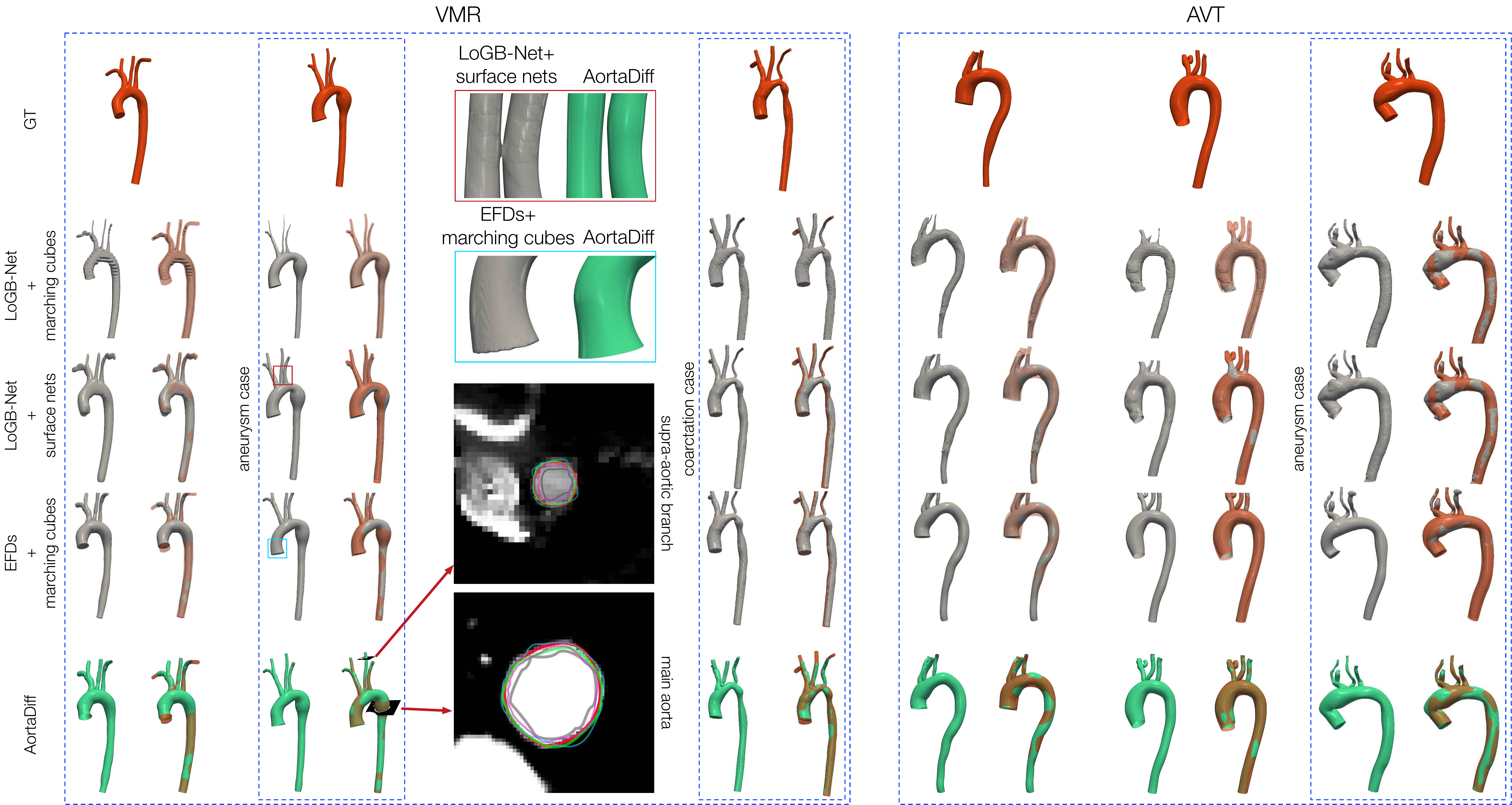} 
    \vspace{-0.10in}
    \caption{\hot{Comparison of mesh construction methods on the selected test cases from the VMR and AVT datasets.
        For each case, we overlay the opaque constructed mesh with the semi-transparent GT mesh to facilitate visual comparison.
        For the aneurysm case from the VMR dataset, we show a zoomed-in comparison of the mesh constructed by EFDs+marching cubes, LoGB-Net+surface nets, and AortaDiff. 
        We also provide a zoomed-in view of the cross-sections of the main aorta and a supra-aortic branch. 
        In these cross-sections, the red, gray, pink, blue, and green curves represent 
        the meshes from GT, 
        LoGB-Net+marching cubes, 
        LoGB-Net+surface nets, 
        EFDs+marching cubes, and 
        AortaDiff, respectively.
        }}
    \label{mesh-results}
\end{figure*}

\vspace{-0.05in}
\subsection{Centerline Generation}

AortaDiff decomposes the aorta into a centerline and its corresponding orthogonal contours, with the centerline serving as the foundation for mesh construction.
Following Section~\ref{sec:cl_rep}, we use a 1D-UNet to process the 1D image representation of the centerline.
Alternatively, a point cloud generation model could be employed since the centerline can be viewed as a 3D point cloud.
To validate our design choice, we compare AortaDiff against three baseline models: {\em diffusion probabilistic model} (DPM)~\cite{LuoH21}, {\em point-voxel diffusion} (PVD)~\cite{PVD}, and Dit-3D~\cite{Dit3D}.
These state-of-the-art baselines are chosen as their implementations are publicly available.
However, because these models are inherently generative, the shape of their generated centerlines varies stochastically.
To ensure a fair comparison, we adapt them into conditional generative models by following AortaDiff's conditioning strategy.
Specifically, we use a \hot{ViT~\cite{ViT}} encoder to extract features from the input volume, align them with the centerline points, and concatenate them with the point cloud input at each diffusion step.
The qualitative evaluation results are presented in Figure~\ref{cl-baselines}.
We can see that AortaDiff is the clear winner across the selected cases from the VMR and AVT datasets.

To assess the similarity between the generated and GT point clouds across different methods, we employ three metrics: {\em chamfer distance} (CD)~\cite{CD}, {\em Hausdorff distance} (HD)~\cite{HD}, and {\em Earth mover's distance} (EMD)~\cite{EMD}.
CD measures the overall quality of the generated point clouds by computing the average nearest-neighbor distance.
HD captures the worst-case discrepancy by quantifying the greatest minimum distance between two sets of points, providing insight into the maximum bias in the generated centerlines.
EMD evaluates the global structural alignment between the generated and reference point clouds.

The quantitative evaluation results across all test cases are presented in Table~\ref{tab:cl-baselines}.
AortaDiff consistently outperforms all baselines. 
On the VMR dataset, AortaDiff achieves a CD of 0.33 mm, an HD of 0.31 mm, and an EMD of 0.17 mm, demonstrating significantly improved accuracy over the baseline models.
Even on the AVT dataset, despite differences in image acquisition (CT modality), AortaDiff maintains superior performance, highlighting its robustness and generalization capability.
These results confirm that AortaDiff produces more accurate and structurally consistent centerlines than baseline models. 
\hot{The centerline diffusion process is presented in Section 2 of the appendix.}

\vspace{-0.075in}
\subsection{Contour Extraction}
\vspace{-0.025in}

The centerline generated in Section~\ref{sec:cl_rep} does not form a perfectly smooth curve, leading to intersections between adjacent cross-sectional planes defined by normal vectors from neighboring centerline points.
These intersections negatively impact contour extraction, potentially compromising the subsequent NURBS mesh construction.
We apply NURBS curve fitting to address this issue, using the original centerline points as control points for a B-spline curve.
A suitable knot vector is defined, and the fitted curve is uniformly resampled to produce a smooth and continuous centerline representation.

Building on the refined centerlines, AortaDiff uses ScribblePrompt to extract aortic contours on each cross-sectional plane of the input volume.
This step is crucial for constructing the aorta mesh using the NURBS method.
We compare AortaDiff with state-of-the-art segmentation approaches, including SAM-based methods—SAM and MedSAM~\cite{MedSAM}—as well as the supervised learning-based \hot{TotalSegmentator~\cite{TotalSegmentator} (TotalSeg) and LoGB-Net~\cite{an2025hierarchical}.}
Multiple prompt formats, such as bounding boxes and points, can guide segmentation for SAM-based models.
Here, we use the generated centerline points as point prompts to facilitate segmentation, ensuring spatial alignment between the extracted contours and the centerline.
In contrast to \hot{TotalSeg} and LoGB-Net's direct volume-to-segmentation approach, which processes the entire CT volume as input, AortaDiff operates on orthogonal cross-sections, leveraging localized structural cues to enhance segmentation accuracy.
To quantitatively evaluate segmentation performance, we employ three metrics: the {\em Dice coefficient} (Dice)~\cite{Dice}, {\em average surface distance} (ASD)~\cite{ASD}, and HD.
The results are summarized in Table~\ref{tab:segmentation-baselines}.

The results show that compared to SAM, MedSAM and AortaDiff significantly enhance segmentation performance, thanks to their training using large-scale medical imaging data, which strengthens their ability to capture anatomical structures.
However, despite utilizing centerline-based prompts, MedSAM still struggles to accurately delineate the aortic region, particularly in the supra-aortic branches.
\hot{LoGB-Net, a fully supervised model specifically designed for aorta segmentation, outperforms both SAM and MedSAM across all metrics. TotalSeg also achieves strong performance on the main aorta but detects only fragmented segments of the supra-aortic branches, as it is a general-purpose model not optimized for these structures. Nevertheless, both LoGB-Net and TotalSeg remain inferior to AortaDiff, likely because they cannot incorporate structural information from centerline point prompts.}
In contrast, AortaDiff benefits from ScribblePrompt's strong prior knowledge gained through large-scale self-supervised pretraining, enabling it to generalize better to unseen vessel structures.

To further evaluate segmentation performance, we visualize the results of different methods in Figure~\ref{segmentation}.
The results indicate that SAM-based methods struggle to accurately segment the aorta, particularly in the supra-aortic branches.
For example, in the third and fourth rows of Figure~\ref{segmentation}, the segmentation results of SAM and MedSAM show significant deviations from the GT in the AVT dataset. In contrast, for the VMR dataset, these methods occasionally produce segmentations that approximate the GT in the supra-aortic branches.
This inconsistency suggests that SAM-based methods lack the robustness needed for aorta segmentation across different datasets, likely due to their reliance on prompts that fail to fully capture the complex anatomical variations.
Meanwhile, \hot{TotalSeg,} LoGB-Net, and AortaDiff demonstrate superior segmentation performance.
However, \hot{TotalSeg and LoGB-Net still fall short of AortaDiff}. \pin{In particular, TotalSeg sometimes detects only fragments of the supra-aortic branches, resulting in discontinuous or incomplete segmentation in these regions.}
Compared to the fully supervised \hot{methods} that directly extract contours for mesh construction,
AortaDiff benefits from ScribblePrompt's pretraining on large-scale medical imaging data and integrates centerline-based prompts, enhancing its robustness and adaptability to anatomical variability.

In Figure~\ref{fig:contours-results}, we visualize the extracted aortic contours on each cross-sectional plane of the input volume.
The results indicate that segmentation inaccuracies often produce highly distorted vessel contours, particularly in the supra-aortic branches.
Notably, when SAM and MedSAM fail to recognize the aorta, they frequently generate excessively large contours that span most of the image, leading to severe segmentation artifacts.
Refer to Figure~\ref{segmentation}.
In the VMR dataset, the cross-sections exhibit relatively low noise, allowing SAM and MedSAM to precisely segment.
However, the AVT dataset contains noisier cross-sections with additional tissues and organs, which adversely affect the segmentation performance of SAM and MedSAM, resulting in suboptimal outcomes.
This explains the irregular and fragmented contours observed in their segmentation outputs.
\hot{TotalSeg exhibits difficulty in detecting the supra-aortic branches, leading to the loss of their corresponding contours.}
In contrast, AortaDiff extracts contours with significantly greater accuracy and consistency across both the main aorta and supra-aortic branches.
To further validate this observation, we conduct a quantitative analysis of contour extraction across different segmentation methods.
As shown in Table~\ref{tab:contour_pc}, AortaDiff consistently produces contours that more closely align with the GT.

\vspace{-0.075in}
\subsection{Aorta Mesh Construction}
\vspace{-0.025in}

We construct the aorta mesh from the extracted contours using the NURBS method.
Other segmentation methods produce suboptimal contours that fail to meet the requirements for NURBS interpolation and perform worse than those generated by AortaDiff.
Therefore, we only compare the NURBS-constructed meshes from AortaDiff with \hot{the marching cubes~\cite{Marchingcubes} and surface nets~\cite{SurfaceNet} generated meshes} from LoGB-Net~\cite{an2025hierarchical}, a state-of-the-art model for aorta segmentation.
\hot{We also compared with marching cubes-generated meshes from the EFDs-based signed distance field~\cite{EFDs}.}
For the VMR dataset, GT meshes are provided, allowing direct comparison.
In contrast, for the AVT dataset, we manually constructed the GT meshes from the segmentation results using SimVascular to ensure anatomical consistency.
The marching cubes-generated meshes are obtained by converting segmentation labels into mesh representations.
To ensure a fair comparison, only basic smoothing techniques are applied to reduce noise and eliminate small holes while preserving overall geometry.
Each supra-aortic branch is constructed separately and assembled into a complete aorta mesh via NURBS.
The quantitative results in Table~\ref{tab:mesh-baselines} demonstrate that AortaDiff outperforms \hot{other methods} in terms of CD, HD, and EMD.

\begin{table}[htbp]
    \caption{\hot{Average CD, HD, and EMD for the constructed aorta meshes on all test cases from the VMR and AVT datasets.}}
    \vspace{-0.1in}
    \centering
    {\fontsize{5pt}{5pt}\selectfont
        \setlength{\tabcolsep}{1.0pt}
        \begin{tabular}{ccccccccc}
                             & \multicolumn{4}{c}{\hot{VMR}} & \multicolumn{4}{c}{\hot{AVT}} \\
            \cmidrule(lr){2-5} \cmidrule(lr){6-9}
            \hot{metric}
                & \hot{\shortstack{LoGB-Net+ \\ marching cubes}}
                & \hot{\shortstack{LoGB-Net+ \\ surface nets}}
                & \hot{\shortstack{EFDs+ \\ marching cubes}}
                & \hot{AortaDiff}
                & \hot{\shortstack{LoGB-Net+ \\ marching cubes}}
                & \hot{\shortstack{LoGB-Net+ \\ surface nets}}
                & \hot{\shortstack{EFDs+ \\ marching cubes}}
                & \hot{AortaDiff} \\ \hline
            \multicolumn{9}{c}{\hot{supra-aortic branches}} \\
            \hot{CD $\downarrow$} 
                & \hot{0.65$\pm$0.09} & \hot{0.62$\pm$0.08} & \hot{0.58$\pm$0.07} & \hot{\textbf{0.31}$\pm$\textbf{0.04}}
                & \hot{0.80$\pm$0.11} & \hot{0.76$\pm$0.10} & \hot{0.72$\pm$0.09} & \hot{\textbf{0.43}$\pm$\textbf{0.06}} \\
            \hot{HD $\downarrow$} 
                & \hot{2.30$\pm$0.27} & \hot{2.25$\pm$0.25} & \hot{2.10$\pm$0.23} & \hot{\textbf{1.30}$\pm$\textbf{0.22}}
                & \hot{2.60$\pm$0.32} & \hot{2.55$\pm$0.30} & \hot{2.40$\pm$0.28} & \hot{\textbf{1.75}$\pm$\textbf{0.45}} \\
            \hot{EMD $\downarrow$} 
                & \hot{1.25$\pm$0.28} & \hot{1.20$\pm$0.26} & \hot{1.12$\pm$0.24} & \hot{\textbf{0.80}$\pm$\textbf{0.20}}
                & \hot{1.35$\pm$0.32} & \hot{1.30$\pm$0.30} & \hot{1.22$\pm$0.28} & \hot{\textbf{0.92}$\pm$\textbf{0.30}} \\
            \hdashline
            \multicolumn{9}{c}{\hot{main aorta}} \\
            \hot{CD $\downarrow$} 
                & \hot{0.55$\pm$0.07} & \hot{0.52$\pm$0.06} & \hot{0.48$\pm$0.06} & \hot{\textbf{0.26}$\pm$\textbf{0.03}}
                & \hot{0.70$\pm$0.09} & \hot{0.67$\pm$0.08} & \hot{0.63$\pm$0.08} & \hot{\textbf{0.36}$\pm$\textbf{0.05}} \\
            \hot{HD $\downarrow$} 
                & \hot{2.10$\pm$0.23} & \hot{2.05$\pm$0.22} & \hot{1.95$\pm$0.20} & \hot{\textbf{1.18}$\pm$\textbf{0.18}}
                & \hot{2.40$\pm$0.28} & \hot{2.35$\pm$0.27} & \hot{2.25$\pm$0.25} & \hot{\textbf{1.58}$\pm$\textbf{0.42}} \\
            \hot{EMD $\downarrow$} 
                & \hot{1.15$\pm$0.23} & \hot{1.12$\pm$0.22} & \hot{1.08$\pm$0.21} & \hot{\textbf{0.72}$\pm$\textbf{0.18}}
                & \hot{1.25$\pm$0.28} & \hot{1.22$\pm$0.27} & \hot{1.18$\pm$0.25} & \hot{\textbf{0.84}$\pm$\textbf{0.28}} \\
            \hdashline
            \multicolumn{9}{c}{\hot{overall}} \\
            \hot{CD $\downarrow$} 
                & \hot{0.60$\pm$0.08} & \hot{0.57$\pm$0.07} & \hot{0.53$\pm$0.07} & \hot{\textbf{0.28}$\pm$\textbf{0.04}}
                & \hot{0.75$\pm$0.10} & \hot{0.72$\pm$0.09} & \hot{0.68$\pm$0.09} & \hot{\textbf{0.39}$\pm$\textbf{0.06}} \\
            \hot{HD $\downarrow$} 
                & \hot{2.20$\pm$0.25} & \hot{2.15$\pm$0.24} & \hot{2.05$\pm$0.22} & \hot{\textbf{1.23}$\pm$\textbf{0.20}}
                & \hot{2.50$\pm$0.30} & \hot{2.45$\pm$0.29} & \hot{2.35$\pm$0.27} & \hot{\textbf{1.66}$\pm$\textbf{0.44}} \\
            \hot{EMD $\downarrow$} 
                & \hot{1.20$\pm$0.25} & \hot{1.17$\pm$0.24} & \hot{1.12$\pm$0.22} & \hot{\textbf{0.76}$\pm$\textbf{0.19}}
                & \hot{1.30$\pm$0.30} & \hot{1.27$\pm$0.29} & \hot{1.22$\pm$0.27} & \hot{\textbf{0.88}$\pm$\textbf{0.29}} \\
        \end{tabular}
    }
    \label{tab:mesh-baselines}
\end{table}

As shown in Figure~\ref{mesh-results}, the NURBS-constructed meshes from AortaDiff exhibit significantly greater smoothness and structural consistency than the marching cubes-constructed meshes from \hot{other methods}.

While LoGB-Net produces segmentation results that closely resemble the GT, its marching cubes-generated meshes suffer from severe artifacts, particularly in the supra-aortic branches. 
\hot{Meshes from LoGB-Net+surface nets show slight improvements in smoothness and structural continuity compared to those from LoGB-Net+marching cubes, but still exhibit noticeable artifacts.}
These artifacts include missing sections, jagged surfaces, and structural discontinuities, especially in regions with complex vessel bifurcations.
Furthermore, the marching cubes method \pin{applied to the LoGB-Net output} often causes shrinkage or even complete disappearance of supra-aortic branches with small radii.
As a result, its accuracy is highly dependent on the resolution of the input volume.
In contrast, AortaDiff demonstrates robustness against such artifacts, generating meshes that are complete, smooth, and topologically consistent.
\hot{Although EFDs leverage smoothed curves to generate a better implicit field, the resulting meshes are still derived from marching cubes and exhibit artifacts and poor smoothness—though less pronounced than those from LoGB-Net—making them unsuitable for CFD simulations.}
Additionally, the evaluation includes \hot{three challenging cases—two aneurysm cases and one coarctation case—from the VMR and AVT datasets}, further validating AortaDiff's capability to construct aorta meshes with complex geometries.
Notably, AortaDiff was trained solely on the VMR dataset, while the AVT dataset was used exclusively for testing.
Despite this, AortaDiff successfully constructs aorta meshes from the AVT dataset, producing smooth, complete surfaces that align with the vascular structures observed in CT images.
The constructed meshes preserve anatomical details while maintaining structural continuity, even for previously unseen datasets.

\begin{figure*}[htb]
    \centering
    \includegraphics[width=1.0\linewidth]{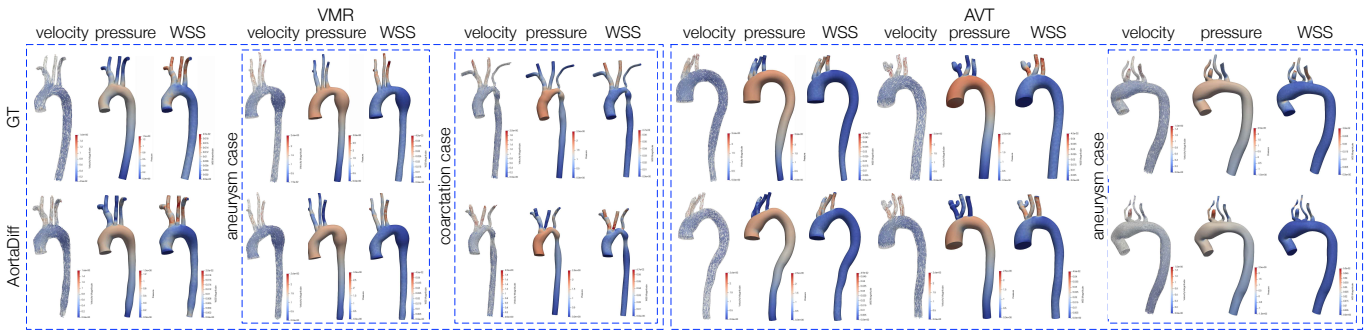} 
    \vspace{-0.25in}
    \caption{\hot{Comparison of hemodynamic simulation results on the selected test cases from the VMR and AVT datasets between aorta meshes constructed using AortaDiff and GT meshes. The visualization includes velocity fields, pressure distributions, and WSS magnitudes.}}
    \label{CFD-results}
\end{figure*}

\vspace{-0.075in}
\subsection{Hemodynamic Simulation}
\vspace{-0.025in}

Using OpenFOAM, we performed hemodynamic simulations on aorta meshes constructed with AortaDiff and compared them to GT meshes. 
\delin{CFD simulations require smooth, watertight surface meshes with anatomically accurate features and sufficiently high resolution to ensure reliable flow field computation, though they do not impose a strict lower bound on angular discretization. This flexibility is a key advantage of our NURBS-based parameterization, which produces simulation-ready geometries that remain smooth and watertight even from sparse or noisy cross-sectional inputs, enabling robust integration with CFD workflows. To ensure simulation reliability, we conducted a mesh independence study, confirming that our chosen resolution—approximately 1 million cells—yields mesh-converged results. Each simulation was run until velocity and pressure residuals fell below $1 \times 10^{-6}$, with continuity errors well controlled at approximately $1 \times 10^{-5}$, ensuring numerical accuracy.}
These simulations capture blood flow velocity, pressure fields, and WSS distributions. 
To ensure a fair comparison, identical boundary conditions were applied to both GT and AortaDiff meshes. 
Meshes constructed from LoGB-Net segmentations were excluded from this experiment due to insufficient geometric fidelity and structural discontinuities, making them unsuitable for CFD analysis.

As shown in Figure~\ref{CFD-results}, the visualization results demonstrate a high degree of consistency in flow dynamics and stress patterns between AortaDiff and GT meshes. 
In particular, the spatial distribution of velocity, pressure, and WSS indicates that AortaDiff constructions accurately preserve key anatomical features critical for precise hemodynamic modeling. 
For both the VMR and AVT datasets, velocity streamlines exhibit similar flow paths and magnitudes, with characteristic acceleration near the ascending aorta and arch, followed by deceleration toward the descending segment. 
Flow direction and lumen confinement remain consistent, with no noticeable divergence or dispersion artifacts. 
Pressure distributions show smooth gradients from the inlet to the distal outlets, with no abrupt discontinuities or localized anomalies in the AortaDiff meshes, closely matching the patterns observed in GT meshes. 
WSS visualizations reveal similar spatial variations between AortaDiff and GT results. 
Elevated WSS values consistently appear on the inner curvature of the arch and branch origins, while lower WSS zones are observed on outer walls and downstream regions. 
Such spatial patterns are preserved across all test cases and datasets, including high-curvature and bifurcation areas. 
These findings suggest that the geometric properties influencing local hemodynamic conditions—such as surface smoothness, curvature continuity, and vessel tapering—are well preserved in AortaDiff constructions. 
As a result, the simulations produce comparable flow, pressure, and WSS characteristics across all anatomical regions evaluated.

\vspace{-0.075in}
\section{Ad-Hoc Expert Feedback}
\vspace{-0.025in}

AortaDiff was tested and evaluated by a CFD and cardiovascular flow expert with over ten years of research experience, who provided the following feedback: ``{\em AortaDiff offers an efficient and accurate method for generating meshes for hemodynamic simulations, a critical component of image-based CFD in clinical diagnosis and treatment planning. Unlike traditional machine learning-based segmentation models, which typically produce binary voxel images, AortaDiff directly generates simulation-ready meshes that can be seamlessly integrated into any downstream fluid solver.
To verify this capability, I used AortaDiff to generate three distinct aortic geometries from three CT images. The resulting geometries closely aligned with the raw images—an impressive outcome given the speed and full automation of the process.}''

The expert further noted ``{\em To assess the feasibility of these meshes for CFD, I employed OpenFOAM to conduct fluid simulations. The simulation solved the steady-state incompressible Navier-Stokes equations using a parabolic velocity profile at the inlet and a fixed pressure at the outlet. The simulation ran successfully, demonstrating a smooth transition from image processing to hemodynamic analysis.
The flow solution revealed a reasonable pressure drop along the streamwise direction, with the highest pressure observed in the ascending aorta and the aortic arch, and the lowest pressure near the descending aorta. As expected, the pressure remained consistently high at junctions between the main aorta and its branches due to vortex formation and flow separation in these geometrically complex regions.
Similarly, higher WSS values were observed near the branching points of the aortic arch, where flow acceleration and bifurcations occur. This suggests a potential risk of wall deterioration, which may contribute to cardiovascular conditions such as \hot{aneurysms and coarctations.}}''

The expert continued ``{\em This experiment confirms that AortaDiff produces high-quality, accurate, and CFD-compatible meshes that integrate seamlessly into CFD workflows. These meshes enable advanced simulations with higher-fidelity settings, such as fluid-structure interactions and Windkessel boundary conditions.
Additionally, since the generated surface is a NURBS surface, alternative CFD approaches—such as isogeometric analysis—can further refine flow hemodynamics, expanding AortaDiff's applicability in computational hemodynamics research. The parameterized NURBS surfaces also facilitate statistical shape analysis when sufficient patient-specific data is available. This enables systematic quantification of anatomical variations, identification of shape-based risk factors, and development of personalized diagnostic and prognostic models.}

Finally, the expert concluded ``{\em In conclusion, I highly recommend AortaDiff to researchers and engineers working on image-based CFD tasks, as it provides accurate, reliable, and highly efficient meshes.}''

\vspace{-0.05in}
\section{Conclusions and Future Work}
\label{sec:conclusions}

We have introduced AortaDiff, a novel framework for generating 3D aorta meshes directly from volumetric CT/MRI data.
Constructing aortic geometries presents several challenges, including the scarcity of annotated data, the heavy reliance on manual intervention, and the absence of a complete end-to-end solution.
To address these issues, AortaDiff employs a volume-guided CDM to generate aortic centerlines—a particularly effective approach given the limited training data available for point cloud generation.
The generated centerline points serve as prompts for semi-automatic segmentation, enabling precise contour extraction from cross-sectional images and significantly reducing manual effort.
We use these contours to construct a smooth 3D mesh via the NURBS method, ensuring an automated pipeline.

\hot{AortaDiff is designed as a modular framework, where each component, such as point cloud generation, contour segmentation, and mesh construction, can be independently improved or replaced to further enhance the overall mesh quality as new methods emerge.} 
Our experiments show that AortaDiff produces high-quality aorta meshes to support hemodynamic simulations comparable to those derived from GT meshes, outperforming conventional marching cubes approaches.
\pin{Furthermore, image-based CFD results generated from our meshes provide clinically valuable insights, including biomarkers derived from WSS and pressure drop analyses in aortic coarctation, supporting both diagnosis and treatment planning. The parametric, NURBS-based model enables flexible hemodynamic recomputation under customized anatomical modifications, enabling in silico experimentation for surgical planning and device design. Unlike traditional voxel- or parameterization-based manual methods, AortaDiff automates segmentation, accelerates mesh generation, and delivers higher-resolution surfaces without increasing manual effort—capabilities that are essential for scalable clinical integration.}

However, AortaDiff still has some remaining limitations. In certain cases, the CDM-generated centerlines are not perfectly centered, though they still reside within the aortic region. While the robustness of the segmentation module mitigates the major impact on performance, this highlights an opportunity for improvement.
\hot{Generating CFD-ready meshes for more complex aortic anatomies, such as aortic dissection, remains an open challenge and will be investigated through structural reasoning and lumen-aware modeling.}

\vspace{-0.05in}
\acknowledgments{This research was supported in part by the University of Notre Dame's Scientific Artificial Intelligence (SAI) Graduate Fellowship, the U.S.\ National Science Foundation through grants IIS-1955395, OAC-2047127, IIS-2101696, OAC-2104158, and IIS-2401144, and the U.S.\ National Institutes of Health through grant 1R01HL177814. Delin An would like to thank Drs.\ Emily Johnson and Donald Brower for their advice on this work through the SAI program. The authors would like to thank the anonymous reviewers for their insightful comments.}

\vspace{-0.05in}
\section*{Appendix}

\setcounter{section}{0}
\setcounter{figure}{0}
\setcounter{table}{0}

\section{Algorithm Description}

As outlined in Algorithm~\ref{alg:aortadiff}, AortaDiff starts with centerline generation through a volume-guided CDM, where Gaussian noise is progressively denoised across timesteps.
The equation that computes $\mathbf{CI}_{t-1}$ involves $\alpha_t$, which balances the noisy centerline and the predicted noise, while $\sigma_t$ scales the Gaussian noise $\mathbf{z}$ added during the reverse diffusion process.
This ensures the final centerline, $\mathbf{CI}_0$, accurately aligns with the anatomical structure.
We use orthogonal slicing planes along the aorta to extract lumen contours via SAM-based ScribblePrompt.
These contours are then resampled, mapped to 3D, and fitted with NURBS to produce a smooth aortic surface construction.

\begin{algorithm}[htb]
    \caption{Constructing 3D aorta mesh}
    \SetAlgoLined
    \label{alg:aortadiff}
    \textbf{Input}: CT/MRI volume $\mathbf{V}$\\
    \textbf{Output}: Constructed aorta mesh\\

    \textbf{Step 1: centerline generation}\\
    Initialize Gaussian noise $\mathbf{L}_t \sim \mathcal{N}(0, \mathbf{I})$, where $\mathbf{I}$ is the identity matrix\\
    Extract feature volume $\mathbf{FV}$ from $\mathbf{V}$ using ViT\\
    \For{each training sample $(\mathbf{V}, \mathbf{CI})$}{
        Sample $t \sim \mathcal{U}(1, T)$\\
        Generate noise $\epsilon \sim \mathcal{N}(0, \mathbf{I})$\\
        Compute noisy centerline $\mathbf{CI}_t \leftarrow \sqrt{\alpha_t}\mathbf{CI} + \sqrt{1 - \alpha_t}\epsilon$\\
        Minimize loss function (Equation 1)\\
    }
    \For{$t = T$ to $0$}{
        Predict noise $\epsilon_t \leftarrow \epsilon_\theta(\mathbf{CI}_t, t, \mathbf{f})$\\
        Compute $\mathbf{CI}_{t-1} \leftarrow \frac{1}{\sqrt{\alpha_t}} \left( \mathbf{CI}_t - \frac{1-\alpha_t}{\sqrt{1-\alpha_t^2}} \epsilon_t \right) + \sigma_t \mathbf{z}$, where $\mathbf{z} \sim \mathcal{N}(0, \mathbf{I})$ if $t > 1$, otherwise $\mathbf{z} = \mathbf{0}$\\
    }
    Obtain centerline points from $\mathbf{CI}_0$\\

    \textbf{Step 2: contour extraction}\\
    \For{each centerline point $p$}{
        Compute tangent vector $\mathbf{t}_p$ and define slicing plane $\Pi_p$\\
        Extract orthogonal slice $\mathbf{S}_p$ from $\mathbf{V}$ (Equation 2)\\
        Segment $\mathbf{S}_p$ via ScribblePrompt to obtain lumen mask\\
        Extract lumen boundary\\
        Resample contour from $\mathcal{C}^{\twod}_p$\\
        Map $\mathcal{C}^{\twod}_p$ back to $\mathcal{C}^{\threed}_p$\\
    }

    \textbf{Step 3: aortic surface construction}\\
    Align adjacent contours to maintain consistency\\
    Fit NURBS surfaces to centerline's contours (Equations 3 and 4)\\
    Repeat \textbf{Steps 1} to \textbf{3} for each supra-aortic branch\\
    Integrate supra-aortic branches with the main aorta\\

    \textbf{Return} constructed aorta mesh\\

\end{algorithm}

\vspace{-0.05in}
\section{Centerline Diffusion Process}

\begin{figure*}[htb]
    \centering
    \includegraphics[width=1.0\linewidth]{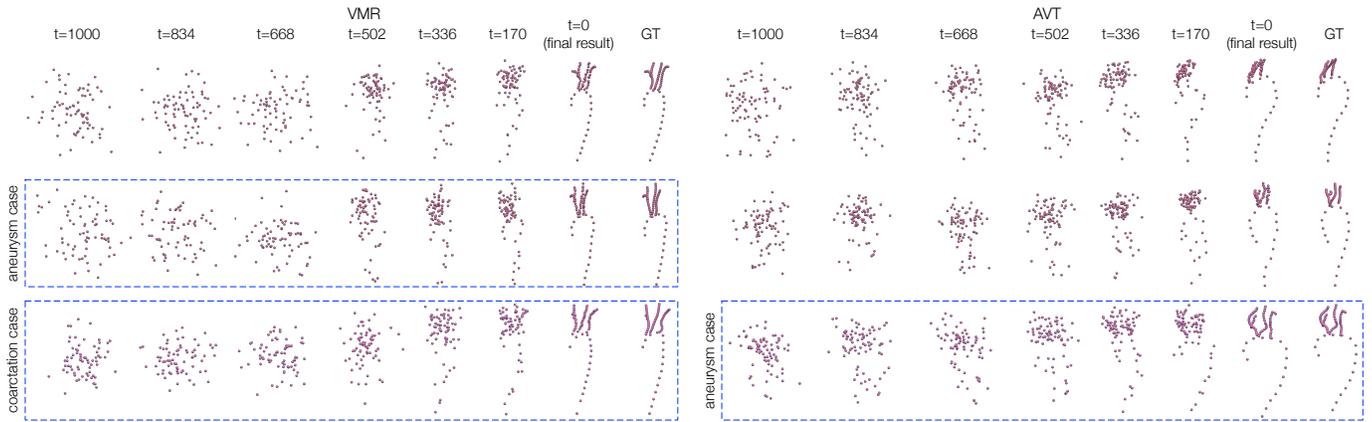} 
    \vspace{-0.25in}
    \caption{\hot{Visualization of the AortaDiff's centerline diffusion process on the selected test cases from the VMR and AVT datasets, illustrating the progressive refinement of centerline points as the diffusion process transitions from Gaussian noise ($t=1000$) to the final prediction ($t=0$).}}
    \label{cl-diffusion}
\end{figure*}

To further evaluate AortaDiff's centerline generation, Figure~\ref{cl-diffusion} visualizes intermediate steps of the diffusion process on the six selected cases from the VMR and AVT datasets. 
The figure illustrates how centerline points progressively emerge from Gaussian noise and are refined under the guidance of features extracted from the input volume. 
These results highlight AortaDiff's ability to handle diverse aortic centerline morphologies. 
Notably, the test set includes two aneurysm cases and one coarctation case, further demonstrating AortaDiff's effectiveness in modeling complex vascular structures.

\vspace{-0.05in}
\section{Additional Results}

\hot{We present additional results, including the remaining one test case from the VMR dataset and \pin{five} test cases from the AVT dataset. 
The results are centerline generation, contour extraction, mesh construction, and hemodynamic simulation, corresponding to Figures 4, 6, 7, and 8 in the paper.}

\begin{figure}[htb]
    \centering
    \includegraphics[height=5.0in]{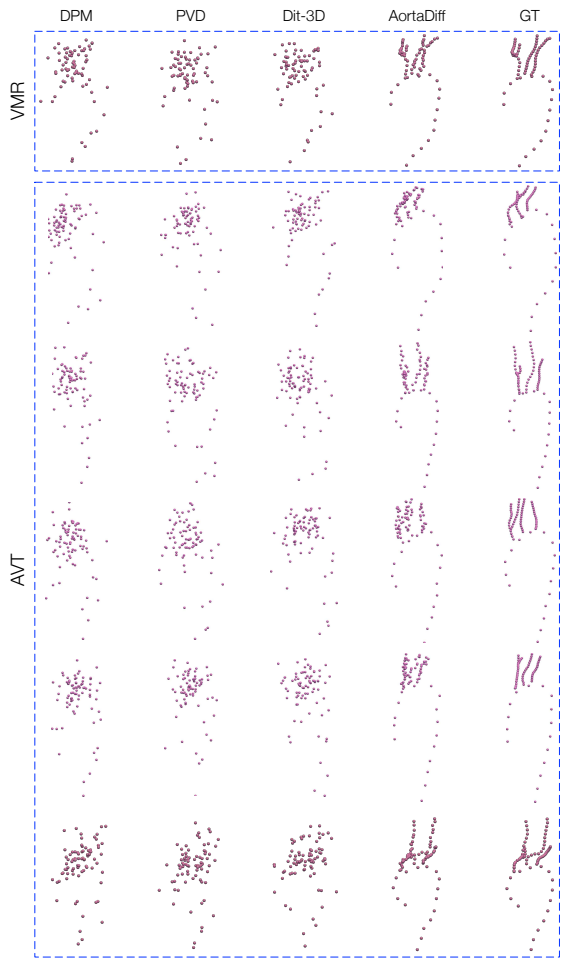}
    \vspace{-0.1in}
    \caption{\hot{Comparison of centerline generation methods on the remaining test cases from the VMR and AVT datasets.}}
    \label{fig:cl}
\end{figure}

{\bf Centerline generation.} 
Figure~\ref{fig:cl} presents the centerline generation results for the remaining test cases, further highlighting the challenges point cloud-based methods face in reliably capturing aortic structures from volumetric inputs under limited data conditions. 
Despite these difficulties, AortaDiff successfully generates anatomically plausible centerlines that accurately reflect the overall vessel morphology, even on the AVT dataset, which differs in source and modality from the training VMR dataset.

\begin{figure}[htb]
    \centering
    \includegraphics[height=5.0in]{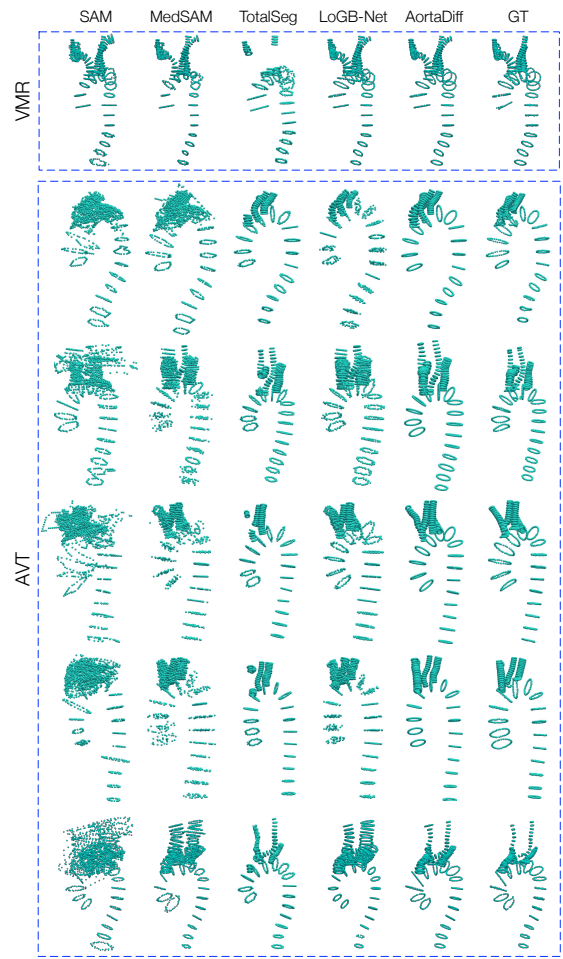}
    \vspace{-0.1in}
    \caption{\hot{Comparison of contours extracted from different segmentation methods on the remaining test cases from the VMR and AVT datasets.}
    }
    \label{fig:contour}
\end{figure}

{\bf Contour extraction.}
Figure~\ref{fig:contour} presents the contour extraction results for the remaining test cases. 
Performance differences between methods are particularly evident in regions with high anatomical variability, such as bifurcations and narrow supra-aortic branches. 
SAM and MedSAM exhibit inconsistent contour quality, often generating fragmented or excessively large regions, likely due to their sensitivity to prompt position and limited adaptability across imaging domains. 
\pin{TotalSeg detects only fragments of the supra-aortic branches, leading to incomplete contour extraction, which is consistent with the results presented in Figure 6 of the paper.}
While LoGB-Net provides more stable segmentation, its results sometimes deviate from vessel lumen boundaries. 
In contrast, AortaDiff consistently produces smooth, well-localized contours by leveraging diffusion-based structural priors and prompt-driven segmentation.

\begin{figure*}[htb]
    \centering
    \includegraphics[width=0.85\linewidth]{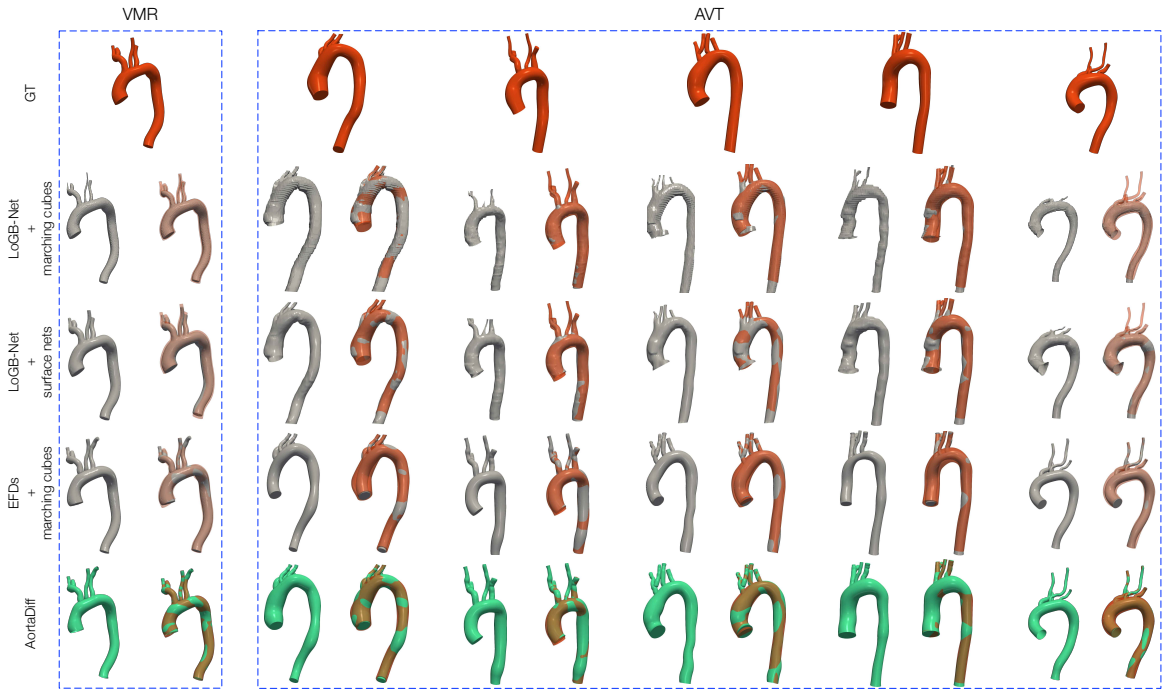}
    \vspace{-0.1in}
    \caption{\hot{
    Comparison of mesh construction methods on the remaining test cases from the VMR and AVT datasets.    
        For each case, we overlay the opaque constructed mesh with the semi-transparent GT mesh to facilitate visual comparison.}
    }
    \label{fig:mesh}
\end{figure*}

{\bf Mesh construction.}
Figure~\ref{fig:mesh} visualizes the mesh construction results for the remaining test cases. 
While LoGB-Net combined with marching cubes can generate coarse surface approximations, the resulting meshes often exhibit aliasing artifacts, surface discontinuities, or missing small branches. 
\pin{EFDs+marching cubes and LoGB-Net+surface nets produce meshes with improved smoothness and topological consistency compared to LoGB-Net+marching cubes, but the results still exhibit artifacts and incorrect topologies, particularly in regions with complex branching structures.}
In contrast, AortaDiff produces smooth, topologically consistent meshes closely aligned with the GT surfaces. 
Its geometry-preserving properties stem from utilizing cross-sectional contours aligned with the centerline trajectory, enabling smooth NURBS surface interpolation. 
This automated, modular pipeline eliminates manual post-processing, significantly reducing the time and effort required for mesh preparation.

{\bf Hemodynamic simulation.}
We perform hemodynamic simulations on the remaining VMR and AVT meshes generated by AortaDiff. 
The visualization results are presented in Figure~\ref{fig:cfd}.
Consistent with the cases presented in the paper, the resulting velocity, pressure, and WSS fields exhibit physiologically meaningful distributions. 
Flow streamlines remain coherent, pressure gradients are smoothly distributed, and WSS concentrations appear in expected regions, such as bifurcation points and inner curvature zones. 
These results further validate that AortaDiff meshes are suitable for CFD analysis, providing a practical alternative to manual mesh generation while maintaining simulation quality. 

\begin{figure*}[htb]
    \centering
    \includegraphics[width=1.0\linewidth]{figs/app-cfd.pdf}
    \vspace{-0.25in}
    \caption{\hot{Comparison of hemodynamic simulation results on the remaining test cases from the VMR and AVT datasets between aorta meshes constructed using AortaDiff and GT meshes. The visualization includes velocity fields, pressure distributions, and WSS magnitudes.}}
    \label{fig:cfd}
\end{figure*}

\vspace{-0.05in}
\section{Parameter Study on Number of Centerline Points}
\begin{figure}[htb]
    \centering
    \includegraphics[width=1.0\linewidth]{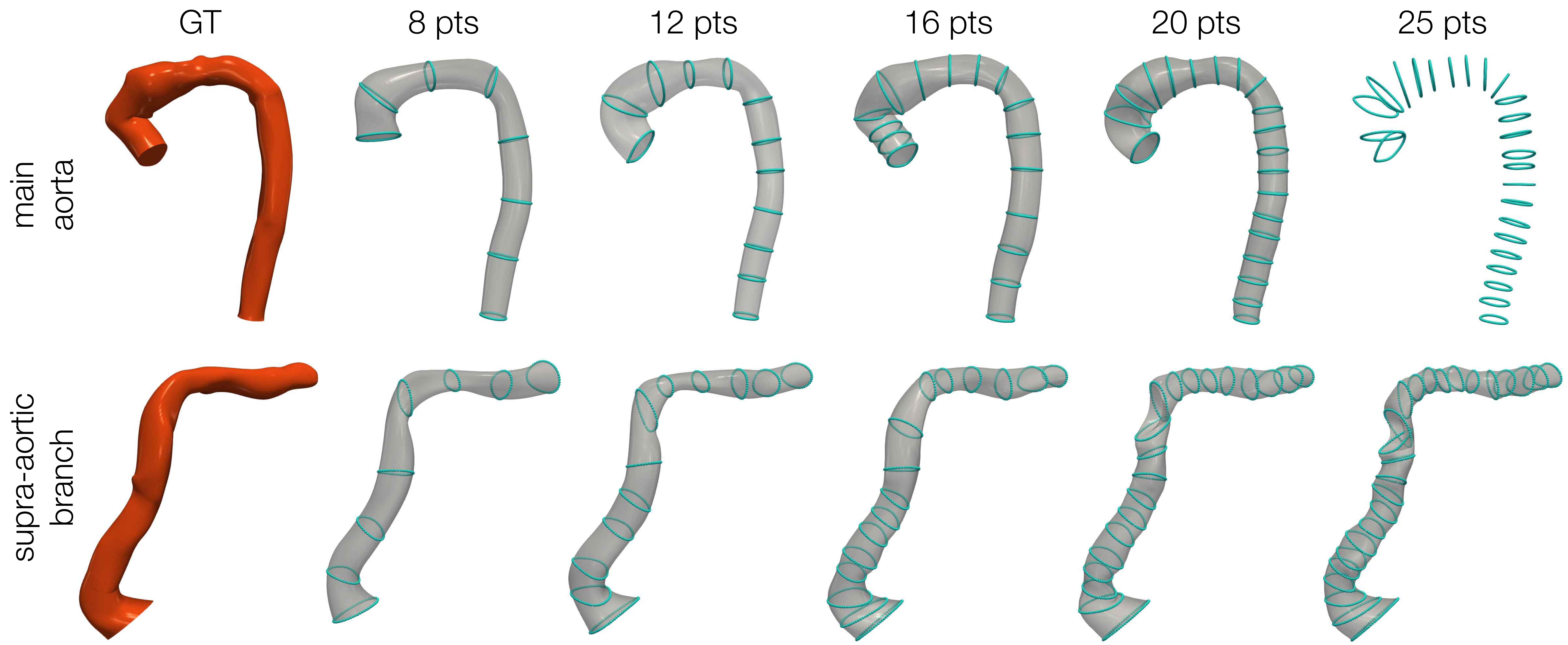}
    \vspace{-0.25in}
    \caption{\hot{Parameter study on the effect of varying the number of centerline points on mesh construction. 
Results are shown for the meshes generated using 8, 12, 16, 20, and 25 centerline points. 
Using 16 points provides the best balance, effectively preserving anatomical detail without introducing redundancy or geometric artifacts.}}
    \label{fig:cl-ablation}
\end{figure}

\hot{
To evaluate the impact of the number of centerline points on the final mesh quality using NURBS-based construction, we conducted a parameter study on all test cases from the VMR and AVT datasets.
The quantitative results are reported in Table~\ref{tab:cl-ablation}. 
A selected example of the main aorta and supra-aortic branch is illustrated in Figure~\ref{fig:cl-ablation}. 
The qualitative results demonstrate that when the number of centerline points is fewer than 16, the reconstructed meshes lose substantial geometric detail, resulting in large errors. 
Conversely, when the point count exceeds 16, overly dense contours—especially in regions with frequent aortic bends—lead to mesh construction failures or artifacts. 
For example, in the main aorta with 25 points, contour interference caused by excessive density resulted in the failure of the NURBS surface fitting. 
These findings indicate that using 16 centerline points achieves the best tradeoff between detail preservation and geometric stability for the AortaDiff's design.
}

\begin{table*}[htb]
    \caption{\hot{Parameter study on the impact of the number of centerline points on final mesh quality.}}
    \vspace{-0.1in}
    \centering
    \resizebox{4.5in}{!}{
        \begin{tabular}{ccccccccccc}
                             & \multicolumn{5}{c}{\hot{VMR}} & \multicolumn{5}{c}{\hot{AVT}} \\
            \cmidrule(lr){2-6} \cmidrule(lr){7-11}
            \hot{metric} & \hot{8 pts} & \hot{12 pts} & \hot{16 pts} & \hot{20 pts} & \hot{25 pts}
                   & \hot{8 pts} & \hot{12 pts} & \hot{16 pts} & \hot{20 pts} & \hot{25 pts} \\ \hline
            \multicolumn{11}{c}{\hot{supra-aortic branches}} \\
            \hot{CD $\downarrow$}  & \hot{0.50$\pm$0.07} & \hot{0.38$\pm$0.06} & \hot{\textbf{0.31}$\pm$\textbf{0.04}} & \hot{0.35$\pm$0.05} & \hot{0.45$\pm$0.07}
                              & \hot{0.60$\pm$0.08} & \hot{0.48$\pm$0.07} & \hot{\textbf{0.43}$\pm$\textbf{0.06}} & \hot{0.45$\pm$0.08} & \hot{0.52$\pm$0.08} \\
            \hot{HD $\downarrow$}  & \hot{2.10$\pm$0.30} & \hot{1.65$\pm$0.25} & \hot{\textbf{1.30}$\pm$\textbf{0.22}} & \hot{1.50$\pm$0.25} & \hot{1.90$\pm$0.30}
                              & \hot{2.40$\pm$0.32} & \hot{1.85$\pm$0.28} & \hot{\textbf{1.75}$\pm$\textbf{0.45}} & \hot{1.77$\pm$0.28} & \hot{2.10$\pm$0.32} \\
            \hot{EMD $\downarrow$} & \hot{1.10$\pm$0.22} & \hot{0.90$\pm$0.20} & \hot{\textbf{0.80}$\pm$\textbf{0.20}} & \hot{0.85$\pm$0.20} & \hot{1.00$\pm$0.22}
                              & \hot{1.25$\pm$0.25} & \hot{1.00$\pm$0.22} & \hot{\textbf{0.92}$\pm$\textbf{0.30}} & \hot{0.92$\pm$0.52} & \hot{1.10$\pm$0.24} \\
            \hdashline
            \multicolumn{11}{c}{\hot{main aorta}} \\
            \hot{CD $\downarrow$}  & \hot{0.40$\pm$0.06} & \hot{0.32$\pm$0.05} & \hot{\textbf{0.26}$\pm$\textbf{0.03}} & \hot{0.29$\pm$0.04} & \hot{0.35$\pm$0.06}
                              & \hot{0.50$\pm$0.07} & \hot{0.40$\pm$0.06} & \hot{\textbf{0.36}$\pm$\textbf{0.05}} & \hot{0.37$\pm$0.05} & \hot{0.42$\pm$0.06} \\
            \hot{HD $\downarrow$}  & \hot{1.70$\pm$0.25} & \hot{1.35$\pm$0.22} & \hot{\textbf{1.18}$\pm$\textbf{0.18}} & \hot{1.25$\pm$0.20} & \hot{1.50$\pm$0.25}
                              & \hot{2.10$\pm$0.28} & \hot{1.65$\pm$0.24} & \hot{\textbf{1.58}$\pm$\textbf{0.42}} & \hot{1.60$\pm$0.24} & \hot{1.80$\pm$0.28} \\
            \hot{EMD $\downarrow$} & \hot{1.00$\pm$0.20} & \hot{0.80$\pm$0.18} & \hot{\textbf{0.72}$\pm$\textbf{0.18}} & \hot{0.78$\pm$0.18} & \hot{0.88$\pm$0.20}
                              & \hot{1.10$\pm$0.23} & \hot{0.90$\pm$0.20} & \hot{\textbf{0.84}$\pm$\textbf{0.28}} & \hot{0.85$\pm$0.20} & \hot{0.96$\pm$0.22} \\
            \hdashline
            \multicolumn{11}{c}{\hot{overall}} \\
            \hot{CD $\downarrow$}  & \hot{0.45$\pm$0.07} & \hot{0.35$\pm$0.06} & \hot{\textbf{0.28}$\pm$\textbf{0.04}} & \hot{0.32$\pm$0.05} & \hot{0.40$\pm$0.07}
                              & \hot{0.55$\pm$0.08} & \hot{0.42$\pm$0.06} & \hot{\textbf{0.39}$\pm$\textbf{0.06}} & \hot{0.39$\pm$0.08} & \hot{0.47$\pm$0.07} \\
            \hot{HD $\downarrow$}  & \hot{1.90$\pm$0.28} & \hot{1.50$\pm$0.24} & \hot{\textbf{1.23}$\pm$\textbf{0.20}} & \hot{1.38$\pm$0.23} & \hot{1.70$\pm$0.28}
                              & \hot{2.25$\pm$0.30} & \hot{1.75$\pm$0.26} & \hot{\textbf{1.66}$\pm$\textbf{0.44}} & \hot{1.68$\pm$0.26} & \hot{1.95$\pm$0.30} \\
            \hot{EMD $\downarrow$} & \hot{1.05$\pm$0.21} & \hot{0.85$\pm$0.19} & \hot{\textbf{0.76}$\pm$\textbf{0.19}} & \hot{0.82$\pm$0.19} & \hot{0.94$\pm$0.21}
                              & \hot{1.18$\pm$0.24} & \hot{0.95$\pm$0.21} & \hot{\textbf{0.88}$\pm$\textbf{0.29}} & \hot{0.89$\pm$0.21} & \hot{1.03$\pm$0.23} \\
        \end{tabular}
    }
    \label{tab:cl-ablation}
\end{table*}

\vspace{-0.05in}
\bibliographystyle{abbrv-doi-hyperref}
\bibliography{template}

\begin{thebibliography}{10}

\bibitem{MONAI}
{{Medical Open Network for AI (MONAI)}}.
\newblock \url{https://monai.io/}.

\bibitem{OpenCV}
{{Open Computer Vision Library (OpenCV)}}.
\newblock \url{https://opencv.org/}.

\bibitem{OpenFOAM}
{{Open Field Operation and Manipulation (OpenFOAM)}}.
\newblock \url{http://www.openfoam.com/}.

\bibitem{an2025hierarchical}
D.~An, P.~Du, P.~Gu, J.-X. Wang, and C.~Wang.
\newblock Hierarchical {LoG Bayesian} neural network for enhanced aorta segmentation.
\newblock In {\em Proceedings of IEEE International Symposium on Biomedical Imaging}, pp. 1--5, 2025. \href{https://doi.org/10.1109/ISBI60581.2025.10980947}
{doi: {{%
10\hspace{.1pt}\discretionary{.}{%
}{.}\hspace{.4pt}1109\discretionary{/}{%
}{/}ISBI60581\hspace{.1pt}\discretionary{.}{%
}{.}\hspace{.4pt}2025\hspace{.1pt}\discretionary{.}{%
}{.}\hspace{.4pt}10980947}}}


\bibitem{AntigaPBERS08}
L.~Antiga, M.~Piccinelli, L.~Botti, B.~Ene{-}Iordache, A.~Remuzzi, and D.~A. Steinman.
\newblock An image-based modeling framework for patient-specific computational hemodynamics.
\newblock {\em Medical \& Biological Engineering \& Computing}, 46(11):1097--1112, 2008. \href{https://doi.org/10.1007/S11517-008-0420-1}
{doi: {{%
10\hspace{.1pt}\discretionary{.}{%
}{.}\hspace{.4pt}1007\discretionary{/}{%
}{/}S11517\discretionary{%
}{-}{-}008\discretionary{%
}{-}{-}0420\discretionary{%
}{-}{-}1}}}


\bibitem{AuerG10}
M.~Auer and T.~C. Gasser.
\newblock Reconstruction and finite element mesh generation of abdominal aortic aneurysms from computerized tomography angiography data with minimal user interactions.
\newblock {\em IEEE Transactions on Medical Imaging}, 29(4):1022--1028, 2010. \href{https://doi.org/10.1109/TMI.2009.2039579}
{doi: {{%
10\hspace{.1pt}\discretionary{.}{%
}{.}\hspace{.4pt}1109\discretionary{/}{%
}{/}TMI\hspace{.1pt}\discretionary{.}{%
}{.}\hspace{.4pt}2009\hspace{.1pt}\discretionary{.}{%
}{.}\hspace{.4pt}2039579}}}


\bibitem{CD}
H.~G. Barrow, J.~M. Tenenbaum, R.~C. Bolles, and H.~C. Wolf.
\newblock Parametric correspondence and chamfer matching: Two new techniques for image matching.
\newblock In {\em Proceedings of International Joint Conference on Artificial Intelligence}, pp. 659--663, 1977. \href{https://doi.org/10.5555/1622943.1622971}
{doi: {{%
10\hspace{.1pt}\discretionary{.}{%
}{.}\hspace{.4pt}5555\discretionary{/}{%
}{/}1622943\hspace{.1pt}\discretionary{.}{%
}{.}\hspace{.4pt}1622971}}}


\bibitem{ICP}
P.~J. Besl and N.~D. McKay.
\newblock A method for registration of {3-D} shapes.
\newblock {\em IEEE Transactions on Pattern Analysis and Machine Intelligence}, 14(2):239--256, 1992. \href{https://doi.org/10.1109/34.121791}
{doi: {{%
10\hspace{.1pt}\discretionary{.}{%
}{.}\hspace{.4pt}1109\discretionary{/}{%
}{/}34\hspace{.1pt}\discretionary{.}{%
}{.}\hspace{.4pt}121791}}}


\bibitem{black2023}
S.~M. Black, C.~Maclean, P.~H. Barrientos, K.~Ritos, and A.~Kazakidi.
\newblock Reconstruction and validation of arterial geometries for computational fluid dynamics using multiple temporal frames of {4D} flow-{MRI} magnitude images.
\newblock {\em Cardiovascular Engineering and Technology}, 14(5):655--676, 2023. \href{https://doi.org/10.1007/s13239-023-00679-x}
{doi: {{%
10\hspace{.1pt}\discretionary{.}{%
}{.}\hspace{.4pt}1007\discretionary{/}{%
}{/}s13239\discretionary{%
}{-}{-}023\discretionary{%
}{-}{-}00679\discretionary{%
}{-}{-}x}}}


\bibitem{Image_Diff}
Y.~Chen, O.~Wang, R.~Zhang, E.~Shechtman, X.~Wang, and M.~Gharbi.
\newblock Image neural field diffusion models.
\newblock In {\em Proceedings of IEEE/CVF Conference on Computer Vision and Pattern Recognition}, pp. 8007--8017, 2024. \href{https://doi.org/10.1109/CVPR52733.2024.00765}
{doi: {{%
10\hspace{.1pt}\discretionary{.}{%
}{.}\hspace{.4pt}1109\discretionary{/}{%
}{/}CVPR52733\hspace{.1pt}\discretionary{.}{%
}{.}\hspace{.4pt}2024\hspace{.1pt}\discretionary{.}{%
}{.}\hspace{.4pt}00765}}}


\bibitem{DanuNVSI19}
M.~Danu, C.~I. Nita, A.~Vizitiu, C.~Suciu, and L.~M. Itu.
\newblock Deep learning based generation of synthetic blood vessel surfaces.
\newblock In {\em Proceedings of International Conference on System Theory, Control and Computing}, pp. 662--667, 2019. \href{https://doi.org/10.1109/ICSTCC.2019.8885576}
{doi: {{%
10\hspace{.1pt}\discretionary{.}{%
}{.}\hspace{.4pt}1109\discretionary{/}{%
}{/}ICSTCC\hspace{.1pt}\discretionary{.}{%
}{.}\hspace{.4pt}2019\hspace{.1pt}\discretionary{.}{%
}{.}\hspace{.4pt}8885576}}}


\bibitem{Dice}
L.~R. Dice.
\newblock Measures of the amount of ecologic association between species.
\newblock {\em Ecology}, 26(3):297--302, 1945. \href{https://doi.org/10.2307/1932409}
{doi: {{%
10\hspace{.1pt}\discretionary{.}{%
}{.}\hspace{.4pt}2307\discretionary{/}{%
}{/}1932409}}}


\bibitem{ViT}
A.~Dosovitskiy, L.~Beyer, A.~Kolesnikov, D.~Weissenborn, X.~Zhai, T.~Unterthiner, M.~Dehghani, M.~Minderer, G.~Heigold, S.~Gelly, et~al.
\newblock An image is worth 16$\times$16 words: Transformers for image recognition at scale.
\newblock In {\em Proceedings of International Conference on Learning Representations}, 2021.

\bibitem{du2025ai}
P.~Du, D.~An, C.~Wang, and J.-X. Wang.
\newblock {AI}-powered automated model construction for patient-specific {CFD} simulations of aortic flows.
\newblock {\em arXiv preprint arXiv:2503.12515}, 2025. \href{https://doi.org/10.48550/arXiv.2503.12515}
{doi: {{%
10\hspace{.1pt}\discretionary{.}{%
}{.}\hspace{.4pt}48550\discretionary{/}{%
}{/}arXiv\hspace{.1pt}\discretionary{.}{%
}{.}\hspace{.4pt}2503\hspace{.1pt}\discretionary{.}{%
}{.}\hspace{.4pt}12515}}}


\bibitem{du2025hugvas}
P.~Du, M.~Xu, X.~Zhu, and J.-X. Wang.
\newblock {HUG-VAS}: A hierarchical {NURBS}-based generative model for aortic geometry synthesis and controllable editing.
\newblock {\em arXiv preprint arXiv:2507.11474}, 2025. \href{https://doi.org/10.48550/arXiv.11474}
{doi: {{%
10\hspace{.1pt}\discretionary{.}{%
}{.}\hspace{.4pt}48550\discretionary{/}{%
}{/}arXiv\hspace{.1pt}\discretionary{.}{%
}{.}\hspace{.4pt}11474}}}


\bibitem{Erler2020}
P.~Erler, P.~Guerrero, S.~Ohrhallinger, M.~Wimmer, and N.~J. Mitra.
\newblock {Points2Surf}: Learning implicit surfaces from point cloud patches.
\newblock {\em arXiv preprint arXiv:2007.10453}, 2020. \href{https://doi.org/10.48550/arXiv.2007.10453}
{doi: {{%
10\hspace{.1pt}\discretionary{.}{%
}{.}\hspace{.4pt}48550\discretionary{/}{%
}{/}arXiv\hspace{.1pt}\discretionary{.}{%
}{.}\hspace{.4pt}2007\hspace{.1pt}\discretionary{.}{%
}{.}\hspace{.4pt}10453}}}


\bibitem{fantazzini20203d}
A.~Fantazzini, M.~Esposito, A.~Finotello, F.~Auricchio, B.~Pane, C.~Basso, G.~Spinella, and M.~Conti.
\newblock {3D} automatic segmentation of aortic computed tomography angiography combining multi-view {2D} convolutional neural networks.
\newblock {\em Cardiovascular Engineering and Technology}, 11:576--586, 2020. \href{https://doi.org/10.1007/s13239-020-00481-z}
{doi: {{%
10\hspace{.1pt}\discretionary{.}{%
}{.}\hspace{.4pt}1007\discretionary{/}{%
}{/}s13239\discretionary{%
}{-}{-}020\discretionary{%
}{-}{-}00481\discretionary{%
}{-}{-}z}}}


\bibitem{SurfaceNet}
S.~F.~F. Gibson.
\newblock Using distance maps for accurate surface representation in sampled volumes.
\newblock In {\em Proceedings of IEEE Symposium on Volume Visualization}, pp. 23--30, 1998. \href{https://doi.org/10.1145/288126.288142}
{doi: {{%
10\hspace{.1pt}\discretionary{.}{%
}{.}\hspace{.4pt}1145\discretionary{/}{%
}{/}288126\hspace{.1pt}\discretionary{.}{%
}{.}\hspace{.4pt}288142}}}


\bibitem{graham2006open}
R.~L. Graham, T.~S. Woodall, and J.~M. Squyres.
\newblock Open {MPI}: A flexible high performance {MPI}.
\newblock In {\em Proceedings of International Conference on Parallel Processing and Applied Mathematics}, pp. 228--239, 2006. \href{https://doi.org/10.1007/11752578_29}
{doi: {{%
10\hspace{.1pt}\discretionary{.}{%
}{.}\hspace{.4pt}1007\discretionary{/}{%
}{/}11752578\_29}}}


\bibitem{Marija2020}
M.~Habijan, I.~Galić, H.~Leventić, K.~Romić, and D.~Babin.
\newblock Abdominal aortic aneurysm segmentation from {CT} images using modified {3D} {U-Net} with deep supervision.
\newblock In {\em Proceedings of International Symposium on Electronics in Marine}, pp. 123--128, 2020. \href{https://doi.org/10.1109/ELMAR49956.2020.9219015}
{doi: {{%
10\hspace{.1pt}\discretionary{.}{%
}{.}\hspace{.4pt}1109\discretionary{/}{%
}{/}ELMAR49956\hspace{.1pt}\discretionary{.}{%
}{.}\hspace{.4pt}2020\hspace{.1pt}\discretionary{.}{%
}{.}\hspace{.4pt}9219015}}}


\bibitem{HahnPSP01}
H.~K. Hahn, B.~Preim, D.~Selle, and H.-O. Peitgen.
\newblock Visualization and interaction techniques for the exploration of vascular structures.
\newblock In {\em Proceedings of IEEE Visualization Conference}, pp. 395--402, 2001. \href{https://doi.org/10.1109/VISUAL.2001.964538}
{doi: {{%
10\hspace{.1pt}\discretionary{.}{%
}{.}\hspace{.4pt}1109\discretionary{/}{%
}{/}VISUAL\hspace{.1pt}\discretionary{.}{%
}{.}\hspace{.4pt}2001\hspace{.1pt}\discretionary{.}{%
}{.}\hspace{.4pt}964538}}}


\bibitem{HD}
F.~Hausdorff.
\newblock {\em Grundz{\"u}ge der {M}engenlehre}.
\newblock Veit \& Comp, Leipzig, 1914.

\bibitem{region-growing}
W.~He, Y.~Cao, Y.~Li, Y.~Miao, W.~Shi, F.~He, F.~Yan, Z.~Jiang, and H.~Zhang.
\newblock A study on {CT} aorta segmentation using vessel enhancement diffusion filter and region growing.
\newblock In {\em Proceedings of International Conference on Natural Computation}, pp. 939--943, 2015. \href{https://doi.org/10.1109/ICNC.2015.7378117}
{doi: {{%
10\hspace{.1pt}\discretionary{.}{%
}{.}\hspace{.4pt}1109\discretionary{/}{%
}{/}ICNC\hspace{.1pt}\discretionary{.}{%
}{.}\hspace{.4pt}2015\hspace{.1pt}\discretionary{.}{%
}{.}\hspace{.4pt}7378117}}}


\bibitem{ASD}
T.~Heimann, B.~van Ginneken, M.~Styner, Y.~Arzhaeva, V.~Aurich, C.~Bauer, A.~Beck, C.~Becker, R.~Beichel, G.~Bekes, et~al.
\newblock Comparison and evaluation of methods for liver segmentation from {CT} datasets.
\newblock {\em IEEE Transactions on Medical Imaging}, 28(8):1251--1265, 2009. \href{https://doi.org/10.1109/TMI.2009.2013851}
{doi: {{%
10\hspace{.1pt}\discretionary{.}{%
}{.}\hspace{.4pt}1109\discretionary{/}{%
}{/}TMI\hspace{.1pt}\discretionary{.}{%
}{.}\hspace{.4pt}2009\hspace{.1pt}\discretionary{.}{%
}{.}\hspace{.4pt}2013851}}}


\bibitem{DDPM}
J.~Ho, A.~Jain, and P.~Abbeel.
\newblock Denoising diffusion probabilistic models.
\newblock In {\em Proceedings of Advances in Neural Information Processing Systems}, 2020.

\bibitem{JinAorta}
Y.~Jin, A.~Pepe, J.~Li, C.~Gsaxner, F.~hua Zhao, K.~L. Pomykala, J.~Kleesiek, A.~F. Frangi, and J.~Egger.
\newblock {AI}-based aortic vessel tree segmentation for cardiovascular diseases treatment: Status quo.
\newblock {\em arXiv preprint arXiv:2108.02998}, 2021. \href{https://doi.org/10.48550/arXiv:2108.02998}
{doi: {{%
10\hspace{.1pt}\discretionary{.}{%
}{.}\hspace{.4pt}48550\discretionary{/}{%
}{/}arXiv\discretionary{:}{%
}{:}2108\hspace{.1pt}\discretionary{.}{%
}{.}\hspace{.4pt}02998}}}


\bibitem{GANLab}
M.~Kahng, N.~Thorat, D.~H.~P. Chau, F.~B. Vi{\'{e}}gas, and M.~Wattenberg.
\newblock {GAN Lab}: Understanding complex deep generative models using interactive visual experimentation.
\newblock {\em IEEE Transactions on Visualization and Computer Graphics}, 25(1):310--320, 2019. \href{https://doi.org/10.1109/TVCG.2018.2864500}
{doi: {{%
10\hspace{.1pt}\discretionary{.}{%
}{.}\hspace{.4pt}1109\discretionary{/}{%
}{/}TVCG\hspace{.1pt}\discretionary{.}{%
}{.}\hspace{.4pt}2018\hspace{.1pt}\discretionary{.}{%
}{.}\hspace{.4pt}2864500}}}


\bibitem{SAM}
A.~Kirillov, E.~Mintun, N.~Ravi, H.~Mao, C.~Rolland, L.~Gustafson, T.~Xiao, S.~Whitehead, A.~C. Berg, W.-Y. Lo, et~al.
\newblock Segment anything.
\newblock In {\em Proceedings of IEEE/CVF International Conference on Computer Vision}, pp. 4015--4026, 2023. \href{https://doi.org/10.1109/ICCV51070.2023.00371}
{doi: {{%
10\hspace{.1pt}\discretionary{.}{%
}{.}\hspace{.4pt}1109\discretionary{/}{%
}{/}ICCV51070\hspace{.1pt}\discretionary{.}{%
}{.}\hspace{.4pt}2023\hspace{.1pt}\discretionary{.}{%
}{.}\hspace{.4pt}00371}}}


\bibitem{KretschmerGPS13}
J.~Kretschmer, C.~Godenschwager, B.~Preim, and M.~Stamminger.
\newblock Interactive patient-specific vascular modeling with sweep surfaces.
\newblock {\em IEEE Transactions on Visualization and Computer Graphics}, 19(12):2828--2837, 2013. \href{https://doi.org/10.1109/TVCG.2013.169}
{doi: {{%
10\hspace{.1pt}\discretionary{.}{%
}{.}\hspace{.4pt}1109\discretionary{/}{%
}{/}TVCG\hspace{.1pt}\discretionary{.}{%
}{.}\hspace{.4pt}2013\hspace{.1pt}\discretionary{.}{%
}{.}\hspace{.4pt}169}}}


\bibitem{KuipersKMB24}
T.~P. Kuipers, P.~Konduri, H.~A. Marquering, and E.~J. Bekkers.
\newblock Generating cerebral vessel trees of acute ischemic stroke patients using conditional set-diffusion.
\newblock In {\em Proceedings of International Conference on Medical Imaging with Deep Learning}, vol. 250, pp. 782--792, 2024.

\bibitem{LawonnGVP016}
K.~Lawonn, S.~Gla{\ss}er, A.~Vilanova, B.~Preim, and T.~Isenberg.
\newblock Occlusion-free blood flow animation with wall thickness visualization.
\newblock {\em IEEE Transactions on Visualization and Computer Graphics}, 22(1):728--737, 2016. \href{https://doi.org/10.1109/TVCG.2015.2467961}
{doi: {{%
10\hspace{.1pt}\discretionary{.}{%
}{.}\hspace{.4pt}1109\discretionary{/}{%
}{/}TVCG\hspace{.1pt}\discretionary{.}{%
}{.}\hspace{.4pt}2015\hspace{.1pt}\discretionary{.}{%
}{.}\hspace{.4pt}2467961}}}


\bibitem{LiLHF21}
R.~Li, X.~Li, P.-A. Heng, and C.-W. Fu.
\newblock Point cloud upsampling via disentangled refinement.
\newblock In {\em Proceedings of IEEE/CVF Conference on Computer Vision and Pattern Recognition}, pp. 344--353, 2021. \href{https://doi.org/10.1109/CVPR46437.2021.00041}
{doi: {{%
10\hspace{.1pt}\discretionary{.}{%
}{.}\hspace{.4pt}1109\discretionary{/}{%
}{/}CVPR46437\hspace{.1pt}\discretionary{.}{%
}{.}\hspace{.4pt}2021\hspace{.1pt}\discretionary{.}{%
}{.}\hspace{.4pt}00041}}}


\bibitem{LinLGG22}
W.~Lin, H.~Liu, L.~Gu, and Z.~Gao.
\newblock A geometry-constrained deformable attention network for aortic segmentation.
\newblock In {\em Proceedings of International Conference on Medical Image Computing and Computer Assisted Interventions}, pp. 287--296, 2022. \href{https://doi.org/10.1007/978-3-031-16443-9_28}
{doi: {{%
10\hspace{.1pt}\discretionary{.}{%
}{.}\hspace{.4pt}1007\discretionary{/}{%
}{/}978\discretionary{%
}{-}{-}3\discretionary{%
}{-}{-}031\discretionary{%
}{-}{-}16443\discretionary{%
}{-}{-}9\_28}}}


\bibitem{Marchingcubes}
W.~E. Lorensen and H.~E. Cline.
\newblock Marching cubes: A high resolution {3D} surface construction algorithm.
\newblock In {\em Proceedings of ACM SIGGRAPH Conference}, pp. 163--169, 1987. \href{https://doi.org/10.1145/37401.37422}
{doi: {{%
10\hspace{.1pt}\discretionary{.}{%
}{.}\hspace{.4pt}1145\discretionary{/}{%
}{/}37401\hspace{.1pt}\discretionary{.}{%
}{.}\hspace{.4pt}37422}}}


\bibitem{LuoH21}
S.~Luo and W.~Hu.
\newblock Diffusion probabilistic models for {3D} point cloud generation.
\newblock In {\em Proceedings of IEEE/CVF Conference on Computer Vision and Pattern Recognition}, pp. 2837--2845, 2021. \href{https://doi.org/10.1109/CVPR46437.2021.00286}
{doi: {{%
10\hspace{.1pt}\discretionary{.}{%
}{.}\hspace{.4pt}1109\discretionary{/}{%
}{/}CVPR46437\hspace{.1pt}\discretionary{.}{%
}{.}\hspace{.4pt}2021\hspace{.1pt}\discretionary{.}{%
}{.}\hspace{.4pt}00286}}}


\bibitem{lyu2023}
J.~Lyu, Y.~Fu, M.~Yang, Y.~Xiong, Q.~Duan, C.~Duan, X.~Wang, X.~Xing, D.~Zhang, J.~Lin, et~al.
\newblock Generative adversarial network-based noncontrast {CT} angiography for aorta and carotid arteries.
\newblock {\em Radiology}, 309(2):e230681, 2023. \href{https://doi.org/10.1148/radiol.230681}
{doi: {{%
10\hspace{.1pt}\discretionary{.}{%
}{.}\hspace{.4pt}1148\discretionary{/}{%
}{/}radiol\hspace{.1pt}\discretionary{.}{%
}{.}\hspace{.4pt}230681}}}


\bibitem{MedSAM}
J.~Ma, Y.~He, F.~Li, L.~Han, C.~You, and B.~Wang.
\newblock Segment anything in medical images.
\newblock {\em Nature Communications}, 15(1):654:1--654:9, 2024. \href{https://doi.org/10.1038/s41467-024-44824-z}
{doi: {{%
10\hspace{.1pt}\discretionary{.}{%
}{.}\hspace{.4pt}1038\discretionary{/}{%
}{/}s41467\discretionary{%
}{-}{-}024\discretionary{%
}{-}{-}44824\discretionary{%
}{-}{-}z}}}


\bibitem{MeuschkeGBWPL19}
M.~Meuschke, T.~G{\"{u}}nther, P.~Berg, R.~Wickenh{\"{o}}fer, B.~Preim, and K.~Lawonn.
\newblock Visual analysis of aneurysm data using statistical graphics.
\newblock {\em IEEE Transactions on Visualization and Computer Graphics}, 25(1):997--1007, 2019. \href{https://doi.org/10.1109/TVCG.2018.2864509}
{doi: {{%
10\hspace{.1pt}\discretionary{.}{%
}{.}\hspace{.4pt}1109\discretionary{/}{%
}{/}TVCG\hspace{.1pt}\discretionary{.}{%
}{.}\hspace{.4pt}2018\hspace{.1pt}\discretionary{.}{%
}{.}\hspace{.4pt}2864509}}}


\bibitem{GUCCI}
M.~Meuschke, U.~Niemann, B.~Behrendt, M.~Gutberlet, B.~Preim, and K.~Lawonn.
\newblock {GUCCI} - guided cardiac cohort investigation of blood flow data.
\newblock {\em IEEE Transactions on Visualization and Computer Graphics}, 29(3):1876--1892, 2023. \href{https://doi.org/10.1109/TVCG.2021.3134083}
{doi: {{%
10\hspace{.1pt}\discretionary{.}{%
}{.}\hspace{.4pt}1109\discretionary{/}{%
}{/}TVCG\hspace{.1pt}\discretionary{.}{%
}{.}\hspace{.4pt}2021\hspace{.1pt}\discretionary{.}{%
}{.}\hspace{.4pt}3134083}}}


\bibitem{MeuschkeOBPL19}
M.~Meuschke, S.~Oeltze{-}Jafra, O.~Beuing, B.~Preim, and K.~Lawonn.
\newblock Classification of blood flow patterns in cerebral aneurysms.
\newblock {\em IEEE Transactions on Visualization and Computer Graphics}, 25(7):2404--2418, 2019. \href{https://doi.org/10.1109/TVCG.2018.2834923}
{doi: {{%
10\hspace{.1pt}\discretionary{.}{%
}{.}\hspace{.4pt}1109\discretionary{/}{%
}{/}TVCG\hspace{.1pt}\discretionary{.}{%
}{.}\hspace{.4pt}2018\hspace{.1pt}\discretionary{.}{%
}{.}\hspace{.4pt}2834923}}}


\bibitem{MeuschkeVBPL17}
M.~Meuschke, S.~Vo{\ss}, O.~Beuing, B.~Preim, and K.~Lawonn.
\newblock Combined visualization of vessel deformation and hemodynamics in cerebral aneurysms.
\newblock {\em IEEE Transactions on Visualization and Computer Graphics}, 23(1):761--770, 2017. \href{https://doi.org/10.1109/TVCG.2016.2598795}
{doi: {{%
10\hspace{.1pt}\discretionary{.}{%
}{.}\hspace{.4pt}1109\discretionary{/}{%
}{/}TVCG\hspace{.1pt}\discretionary{.}{%
}{.}\hspace{.4pt}2016\hspace{.1pt}\discretionary{.}{%
}{.}\hspace{.4pt}2598795}}}


\bibitem{EFDs}
G.~Mistelbauer, C.~R{\"{o}}ssl, K.~B{\"{a}}umler, B.~Preim, and D.~Fleischmann.
\newblock Implicit modeling of patient-specific aortic dissections with elliptic {Fourier} descriptors.
\newblock {\em Computer Graphics Forum}, 40(3):423--434, 2021. \href{https://doi.org/10.1111/CGF.14318}
{doi: {{%
10\hspace{.1pt}\discretionary{.}{%
}{.}\hspace{.4pt}1111\discretionary{/}{%
}{/}CGF\hspace{.1pt}\discretionary{.}{%
}{.}\hspace{.4pt}14318}}}


\bibitem{Dit3D}
S.~Mo, E.~Xie, R.~Chu, L.~Hong, M.~Nie{\ss}ner, and Z.~Li.
\newblock {DiT-3D}: Exploring plain diffusion transformers for {3D} shape generation.
\newblock In {\em Proceedings of Advances in Neural Information Processing Systems}, 2023.

\bibitem{moccia2018blood}
S.~Moccia, E.~D. Momi, S.~E. Hadji, and L.~S. Mattos.
\newblock Blood vessel segmentation algorithms - review of methods, datasets and evaluation metrics.
\newblock {\em Computer Methods and Programs in Biomedicine}, 158:71--91, 2018. \href{https://doi.org/10.1016/j.cmpb.2018.02.001}
{doi: {{%
10\hspace{.1pt}\discretionary{.}{%
}{.}\hspace{.4pt}1016\discretionary{/}{%
}{/}j\hspace{.1pt}\discretionary{.}{%
}{.}\hspace{.4pt}cmpb\hspace{.1pt}\discretionary{.}{%
}{.}\hspace{.4pt}2018\hspace{.1pt}\discretionary{.}{%
}{.}\hspace{.4pt}02\hspace{.1pt}\discretionary{.}{%
}{.}\hspace{.4pt}001}}}


\bibitem{Montalt2022}
J.~Montalt-Tordera, E.~Pajaziti, R.~Jones, E.~Sauvage, R.~Puranik, A.~A.~V. Singh, C.~Capelli, J.~Steeden, S.~Schievano, and V.~Muthurangu.
\newblock Automatic segmentation of the great arteries for computational hemodynamic assessment.
\newblock {\em Journal of Cardiovascular Magnetic Resonance}, 24(1):57, 2022. \href{https://doi.org/10.1186/s12968-022-00891-z}
{doi: {{%
10\hspace{.1pt}\discretionary{.}{%
}{.}\hspace{.4pt}1186\discretionary{/}{%
}{/}s12968\discretionary{%
}{-}{-}022\discretionary{%
}{-}{-}00891\discretionary{%
}{-}{-}z}}}


\bibitem{MyronenkoYHX23}
A.~Myronenko, D.~Yang, Y.~He, and D.~Xu.
\newblock Aorta segmentation from {3D} {CT} in {MICCAI} {SEG.A.} 2023 challenge.
\newblock In {\em Proceedings of International Conference on Medical Image Computing and Computer Assisted Interventions}, pp. 13--18, 2023. \href{https://doi.org/10.1007/978-3-031-53241-2_2}
{doi: {{%
10\hspace{.1pt}\discretionary{.}{%
}{.}\hspace{.4pt}1007\discretionary{/}{%
}{/}978\discretionary{%
}{-}{-}3\discretionary{%
}{-}{-}031\discretionary{%
}{-}{-}53241\discretionary{%
}{-}{-}2\_2}}}


\bibitem{NicholD21}
A.~Q. Nichol and P.~Dhariwal.
\newblock Improved denoising diffusion probabilistic models.
\newblock In {\em Proceedings of International Conference on Machine Learning}, pp. 8162--8171, 2021.

\bibitem{Oeltze-JafraCJP16}
S.~Oeltze{-}Jafra, J.~R. Cebral, G.~Janiga, and B.~Preim.
\newblock Cluster analysis of vortical flow in simulations of cerebral aneurysm hemodynamics.
\newblock {\em IEEE Transactions on Visualization and Computer Graphics}, 22(1):757--766, 2016. \href{https://doi.org/10.1109/TVCG.2015.2467203}
{doi: {{%
10\hspace{.1pt}\discretionary{.}{%
}{.}\hspace{.4pt}1109\discretionary{/}{%
}{/}TVCG\hspace{.1pt}\discretionary{.}{%
}{.}\hspace{.4pt}2015\hspace{.1pt}\discretionary{.}{%
}{.}\hspace{.4pt}2467203}}}


\bibitem{patankar1983calculation}
S.~V. Patankar and D.~B. Spalding.
\newblock A calculation procedure for heat, mass and momentum transfer in three-dimensional parabolic flows.
\newblock In {\em Numerical Prediction of Flow, Heat Transfer, Turbulence and Combustion}, chap.~5, pp. 54--73. Elsevier, 1983. \href{https://doi.org/10.1016/B978-0-08-030937-8.50013-1}
{doi: {{%
10\hspace{.1pt}\discretionary{.}{%
}{.}\hspace{.4pt}1016\discretionary{/}{%
}{/}B978\discretionary{%
}{-}{-}0\discretionary{%
}{-}{-}08\discretionary{%
}{-}{-}030937\discretionary{%
}{-}{-}8\hspace{.1pt}\discretionary{.}{%
}{.}\hspace{.4pt}50013\discretionary{%
}{-}{-}1}}}


\bibitem{piegl1996nurbs}
L.~Piegl and W.~Tiller.
\newblock {\em The NURBS Book (Monographs in Visual Communication)}.
\newblock Springer, second ed., 1996.

\bibitem{Radl2022}
L.~Radl, Y.~Jin, A.~Pepe, J.~Li, C.~Gsaxner, F.-H. Zhao, and J.~Egger.
\newblock {AVT}: Multicenter aortic vessel tree {CTA} dataset collection with ground truth segmentation masks.
\newblock {\em Data in Brief}, 40:107801, 2022. \href{https://doi.org/10.1016/j.dib.2022.107801}
{doi: {{%
10\hspace{.1pt}\discretionary{.}{%
}{.}\hspace{.4pt}1016\discretionary{/}{%
}{/}j\hspace{.1pt}\discretionary{.}{%
}{.}\hspace{.4pt}dib\hspace{.1pt}\discretionary{.}{%
}{.}\hspace{.4pt}2022\hspace{.1pt}\discretionary{.}{%
}{.}\hspace{.4pt}107801}}}


\bibitem{U-Net}
O.~Ronneberger, P.~Fischer, and T.~Brox.
\newblock {U-Net}: Convolutional networks for biomedical image segmentation.
\newblock In {\em Proceedings of International Conference on Medical Image Computing and Computer-Assisted Intervention}, pp. 234--241, 2015. \href{https://doi.org/10.1007/978-3-319-24574-4_28}
{doi: {{%
10\hspace{.1pt}\discretionary{.}{%
}{.}\hspace{.4pt}1007\discretionary{/}{%
}{/}978\discretionary{%
}{-}{-}3\discretionary{%
}{-}{-}319\discretionary{%
}{-}{-}24574\discretionary{%
}{-}{-}4\_28}}}


\bibitem{EMD}
Y.~Rubner, C.~Tomasi, and L.~J. Guibas.
\newblock A metric for distributions with applications to image databases.
\newblock In {\em Proceedings of IEEE/CVF International Conference on Computer Vision}, pp. 59--66, 1998. \href{https://doi.org/10.1109/iccv.1998.710701}
{doi: {{%
10\hspace{.1pt}\discretionary{.}{%
}{.}\hspace{.4pt}1109\discretionary{/}{%
}{/}iccv\hspace{.1pt}\discretionary{.}{%
}{.}\hspace{.4pt}1998\hspace{.1pt}\discretionary{.}{%
}{.}\hspace{.4pt}710701}}}


\bibitem{liverVessels}
D.~Selle, B.~Preim, A.~Schenk, and H.~Peitgen.
\newblock Analysis of vasculature for liver surgical planning.
\newblock {\em IEEE Transactions on Medical Imaging}, 21(11):1344--1357, 2002. \href{https://doi.org/10.1109/TMI.2002.801166}
{doi: {{%
10\hspace{.1pt}\discretionary{.}{%
}{.}\hspace{.4pt}1109\discretionary{/}{%
}{/}TMI\hspace{.1pt}\discretionary{.}{%
}{.}\hspace{.4pt}2002\hspace{.1pt}\discretionary{.}{%
}{.}\hspace{.4pt}801166}}}


\bibitem{UNetR++}
A.~M. Shaker, M.~Maaz, H.~A. Rasheed, S.~H. Khan, M.~Yang, and F.~S. Khan.
\newblock {UNETR++}: Delving into efficient and accurate {3D} medical image segmentation.
\newblock {\em IEEE Transactions on Medical Imaging}, 43(9):3377--3390, 2024. \href{https://doi.org/10.1109/TMI.2024.3398728}
{doi: {{%
10\hspace{.1pt}\discretionary{.}{%
}{.}\hspace{.4pt}1109\discretionary{/}{%
}{/}TMI\hspace{.1pt}\discretionary{.}{%
}{.}\hspace{.4pt}2024\hspace{.1pt}\discretionary{.}{%
}{.}\hspace{.4pt}3398728}}}


\bibitem{BIVDiff}
F.~Shi, J.~Gu, H.~Xu, S.~Xu, W.~Zhang, and L.~Wang.
\newblock {BIVDiff}: A training-free framework for general-purpose video synthesis via bridging image and video diffusion models.
\newblock In {\em Proceedings of IEEE/CVF Conference on Computer Vision and Pattern Recognition}, pp. 7393--7402, 2024. \href{https://doi.org/10.1109/CVPR52733.2024.00706}
{doi: {{%
10\hspace{.1pt}\discretionary{.}{%
}{.}\hspace{.4pt}1109\discretionary{/}{%
}{/}CVPR52733\hspace{.1pt}\discretionary{.}{%
}{.}\hspace{.4pt}2024\hspace{.1pt}\discretionary{.}{%
}{.}\hspace{.4pt}00706}}}


\bibitem{CDM}
V.~Singh, S.~Jandial, A.~Chopra, S.~Ramesh, B.~Krishnamurthy, and V.~N. Balasubramanian.
\newblock On conditioning the input noise for controlled image generation with diffusion models.
\newblock {\em arXiv preprint arXiv:2205.03859}, 2022. \href{https://doi.org/10.48550/arXiv.2205.03859}
{doi: {{%
10\hspace{.1pt}\discretionary{.}{%
}{.}\hspace{.4pt}48550\discretionary{/}{%
}{/}arXiv\hspace{.1pt}\discretionary{.}{%
}{.}\hspace{.4pt}2205\hspace{.1pt}\discretionary{.}{%
}{.}\hspace{.4pt}03859}}}


\bibitem{Aneurysm}
A.~Swillens, L.~Lanoye, J.~D. Backer, N.~Stergiopulos, P.~R. Verdonck, F.~Vermassen, and P.~Segers.
\newblock Effect of an abdominal aortic aneurysm on wave reflection in the aorta.
\newblock {\em IEEE Transactions on Biomedical Engineering}, 55(5):1602--1611, 2008. \href{https://doi.org/10.1109/TBME.2007.913994}
{doi: {{%
10\hspace{.1pt}\discretionary{.}{%
}{.}\hspace{.4pt}1109\discretionary{/}{%
}{/}TBME\hspace{.1pt}\discretionary{.}{%
}{.}\hspace{.4pt}2007\hspace{.1pt}\discretionary{.}{%
}{.}\hspace{.4pt}913994}}}


\bibitem{TaoHQWJSZY16}
J.~Tao, X.~Huang, F.~Qiu, C.~Wang, J.~Jiang, C.~Shene, Y.~Zhao, and D.~Yu.
\newblock {VesselMap}: A web interface to explore multivariate vascular data.
\newblock {\em Computers \& Graphics}, 59:79--92, 2016. \href{https://doi.org/10.1016/J.CAG.2016.05.024}
{doi: {{%
10\hspace{.1pt}\discretionary{.}{%
}{.}\hspace{.4pt}1016\discretionary{/}{%
}{/}J\hspace{.1pt}\discretionary{.}{%
}{.}\hspace{.4pt}CAG\hspace{.1pt}\discretionary{.}{%
}{.}\hspace{.4pt}2016\hspace{.1pt}\discretionary{.}{%
}{.}\hspace{.4pt}05\hspace{.1pt}\discretionary{.}{%
}{.}\hspace{.4pt}024}}}


\bibitem{updegrove2017simvascular}
A.~Updegrove, N.~M. Wilson, J.~Merkow, H.~Lan, A.~L. Marsden, and S.~C. Shadden.
\newblock {SimVascular}: An open source pipeline for cardiovascular simulation.
\newblock {\em Annals of Biomedical Engineering}, 45(3):525--541, 2017.

\bibitem{VagenasGM23}
T.~P. Vagenas, K.~Georgas, and G.~K. Matsopoulos.
\newblock Deep learning-based segmentation and mesh reconstruction of the aortic vessel tree from {CTA} images.
\newblock In {\em Proceedings of International Conference on Medical Image Computing and Computer Assisted Interventions}, pp. 80--94, 2023. \href{https://doi.org/10.1007/978-3-031-53241-2_7}
{doi: {{%
10\hspace{.1pt}\discretionary{.}{%
}{.}\hspace{.4pt}1007\discretionary{/}{%
}{/}978\discretionary{%
}{-}{-}3\discretionary{%
}{-}{-}031\discretionary{%
}{-}{-}53241\discretionary{%
}{-}{-}2\_7}}}


\bibitem{ValetteCP08}
S.~Valette, J.~Chassery, and R.~Prost.
\newblock Generic remeshing of {3D} triangular meshes with metric-dependent discrete {Voronoi} diagrams.
\newblock {\em IEEE Transactions on Visualization and Computer Graphics}, 14(2):369--381, 2008. \href{https://doi.org/10.1109/TVCG.2007.70430}
{doi: {{%
10\hspace{.1pt}\discretionary{.}{%
}{.}\hspace{.4pt}1109\discretionary{/}{%
}{/}TVCG\hspace{.1pt}\discretionary{.}{%
}{.}\hspace{.4pt}2007\hspace{.1pt}\discretionary{.}{%
}{.}\hspace{.4pt}70430}}}


\bibitem{TotalSegmentator}
J.~Wasserthal, H.-C. Breit, M.~T. Meyer, M.~Pradella, D.~Hinck, A.~W. Sauter, T.~Heye, D.~T. Boll, J.~Cyriac, S.~Yang, et~al.
\newblock {TotalSegmentator}: Robust segmentation of 104 anatomic structures in {CT} images.
\newblock {\em Radiology: Artificial Intelligence}, 5(5):e230024, 2023. \href{https://doi.org/10.1148/ryai.230024}
{doi: {{%
10\hspace{.1pt}\discretionary{.}{%
}{.}\hspace{.4pt}1148\discretionary{/}{%
}{/}ryai\hspace{.1pt}\discretionary{.}{%
}{.}\hspace{.4pt}230024}}}


\bibitem{Wilson2013}
N.~M. Wilson, A.~K. Ortiz, and A.~B. Johnson.
\newblock The vascular model repository: A public resource of medical imaging data and blood flow simulation results.
\newblock {\em Journal of Medical Devices}, 7(4):0409231, 2013. \href{https://doi.org/10.1115/1.4025983}
{doi: {{%
10\hspace{.1pt}\discretionary{.}{%
}{.}\hspace{.4pt}1115\discretionary{/}{%
}{/}1\hspace{.1pt}\discretionary{.}{%
}{.}\hspace{.4pt}4025983}}}


\bibitem{BloodVesselGAN}
J.~M. Wolterink, T.~Leiner, and I.~Isgum.
\newblock Blood vessel geometry synthesis using generative adversarial networks.
\newblock {\em arXiv preprint arXiv:1804.04381}, 2018. \href{https://doi.org/10.48550/arXiv:1804.04381}
{doi: {{%
10\hspace{.1pt}\discretionary{.}{%
}{.}\hspace{.4pt}48550\discretionary{/}{%
}{/}arXiv\discretionary{:}{%
}{:}1804\hspace{.1pt}\discretionary{.}{%
}{.}\hspace{.4pt}04381}}}


\bibitem{ScribblePrompt}
H.~E. Wong, M.~Rakic, J.~V. Guttag, and A.~V. Dalca.
\newblock {ScribblePrompt}: Fast and flexible interactive segmentation for any biomedical image.
\newblock In {\em Proceedings of European Conference on Computer Vision}, pp. 207--229, 2024. \href{https://doi.org/10.1007/978-3-031-73661-2_12}
{doi: {{%
10\hspace{.1pt}\discretionary{.}{%
}{.}\hspace{.4pt}1007\discretionary{/}{%
}{/}978\discretionary{%
}{-}{-}3\discretionary{%
}{-}{-}031\discretionary{%
}{-}{-}73661\discretionary{%
}{-}{-}2\_12}}}


\bibitem{Wu23}
L.~Wu, D.~Wang, C.~Gong, X.~Liu, Y.~Xiong, R.~Ranjan, R.~Krishnamoorthi, V.~Chandra, and Q.~Liu.
\newblock Fast point cloud generation with straight flows.
\newblock In {\em Proceedings of IEEE/CVF Conference on Computer Vision and Pattern Recognition}, pp. 9445--9454, 2023. \href{https://doi.org/10.1109/CVPR52729.2023.00911}
{doi: {{%
10\hspace{.1pt}\discretionary{.}{%
}{.}\hspace{.4pt}1109\discretionary{/}{%
}{/}CVPR52729\hspace{.1pt}\discretionary{.}{%
}{.}\hspace{.4pt}2023\hspace{.1pt}\discretionary{.}{%
}{.}\hspace{.4pt}00911}}}


\bibitem{xiong2022}
X.~Xiong, Y.~Ding, C.~Sun, Z.~Zhang, X.~Guan, T.~Zhang, H.~Chen, H.~Liu, Z.~Cheng, L.~Zhao, et~al.
\newblock A cascaded multi-task generative framework for detecting aortic dissection on {3-D} non-contrast-enhanced computed tomography.
\newblock {\em IEEE Journal of Biomedical and Health Informatics}, 26(10):5177--5188, 2022. \href{https://doi.org/10.1109/JBHI.2022.3190293}
{doi: {{%
10\hspace{.1pt}\discretionary{.}{%
}{.}\hspace{.4pt}1109\discretionary{/}{%
}{/}JBHI\hspace{.1pt}\discretionary{.}{%
}{.}\hspace{.4pt}2022\hspace{.1pt}\discretionary{.}{%
}{.}\hspace{.4pt}3190293}}}


\bibitem{Ye_2023_ICCV}
M.~Ye, D.~Yang, M.~Kanski, L.~Axel, and D.~Metaxas.
\newblock Neural deformable models for {3D} bi-ventricular heart shape reconstruction and modeling from {2D} sparse cardiac magnetic resonance imaging.
\newblock In {\em Proceedings of IEEE/CVF International Conference on Computer Vision}, pp. 14247--14256, 2023. \href{https://doi.org/10.1109/ICCV51070.2023.01310}
{doi: {{%
10\hspace{.1pt}\discretionary{.}{%
}{.}\hspace{.4pt}1109\discretionary{/}{%
}{/}ICCV51070\hspace{.1pt}\discretionary{.}{%
}{.}\hspace{.4pt}2023\hspace{.1pt}\discretionary{.}{%
}{.}\hspace{.4pt}01310}}}


\bibitem{Text-to-Image}
Y.~Zeng, V.~M. Patel, H.~Wang, X.~Huang, T.-C. Wang, M.-Y. Liu, and Y.~Balaji.
\newblock {JeDi}: Joint-image diffusion models for finetuning-free personalized text-to-image generation.
\newblock In {\em Proceedings of IEEE/CVF Conference on Computer Vision and Pattern Recognition}, pp. 6786--6795, 2024. \href{https://doi.org/10.1109/CVPR52733.2024.00648}
{doi: {{%
10\hspace{.1pt}\discretionary{.}{%
}{.}\hspace{.4pt}1109\discretionary{/}{%
}{/}CVPR52733\hspace{.1pt}\discretionary{.}{%
}{.}\hspace{.4pt}2024\hspace{.1pt}\discretionary{.}{%
}{.}\hspace{.4pt}00648}}}


\bibitem{ZHANG2025106991}
B.~Zhang, Z.~Lai, S.~Liu, X.~Xie, X.~Zhou, Z.~Hou, X.~Ma, B.~Liu, K.~Li, and M.~Song.
\newblock {SDLU-Net}: A similarity-based dynamic linking network for the automated segmentation of abdominal aorta aneurysms and branching vessels.
\newblock {\em Biomedical Signal Processing and Control}, 100:106991, 2025. \href{https://doi.org/10.1016/j.bspc.2024.106991}
{doi: {{%
10\hspace{.1pt}\discretionary{.}{%
}{.}\hspace{.4pt}1016\discretionary{/}{%
}{/}j\hspace{.1pt}\discretionary{.}{%
}{.}\hspace{.4pt}bspc\hspace{.1pt}\discretionary{.}{%
}{.}\hspace{.4pt}2024\hspace{.1pt}\discretionary{.}{%
}{.}\hspace{.4pt}106991}}}


\bibitem{PVD}
L.~Zhou, Y.~Du, and J.~Wu.
\newblock {3D} shape generation and completion through point-voxel diffusion.
\newblock In {\em Proceedings of IEEE/CVF International Conference on Computer Vision}, pp. 5806--5815, 2021. \href{https://doi.org/10.1109/ICCV48922.2021.00577}
{doi: {{%
10\hspace{.1pt}\discretionary{.}{%
}{.}\hspace{.4pt}1109\discretionary{/}{%
}{/}ICCV48922\hspace{.1pt}\discretionary{.}{%
}{.}\hspace{.4pt}2021\hspace{.1pt}\discretionary{.}{%
}{.}\hspace{.4pt}00577}}}


\bibitem{DenseUNet}
Y.~Zhou, H.~Chang, X.~Lu, and Y.~Lu.
\newblock {DenseUNet}: Improved image classification method using standard convolution and dense transposed convolution.
\newblock {\em Knowledge-Based Systems}, 254:109658, 2022. \href{https://doi.org/10.1016/j.knosys.2022.109658}
{doi: {{%
10\hspace{.1pt}\discretionary{.}{%
}{.}\hspace{.4pt}1016\discretionary{/}{%
}{/}j\hspace{.1pt}\discretionary{.}{%
}{.}\hspace{.4pt}knosys\hspace{.1pt}\discretionary{.}{%
}{.}\hspace{.4pt}2022\hspace{.1pt}\discretionary{.}{%
}{.}\hspace{.4pt}109658}}}


\end{thebibliography}

\end{document}